\PassOptionsToPackage{table}{xcolor}

\documentclass{article} 
\usepackage{iclr2027_conference,times}

\iclrfinalcopy
\makeatletter
\def\@oddhead{}
\def\@evenhead{}
\makeatother

\usepackage{amsmath,amsfonts,bm}

\def\eqref#1{equation~\ref{#1}}

\def\1{\bm{1}}

\DeclareMathAlphabet{\mathsfit}{\encodingdefault}{\sfdefault}{m}{sl}
\SetMathAlphabet{\mathsfit}{bold}{\encodingdefault}{\sfdefault}{bx}{n}

\usepackage{booktabs}
\usepackage{multirow}
\usepackage{graphicx}
\usepackage{array}
\usepackage{makecell}

\usepackage[table]{xcolor}

\definecolor{rowshade}{RGB}{230,238,247}
\usepackage{wrapfig}

\usepackage{xspace}
\newcommand{\ours}{JARVIS\xspace}
\usepackage{booktabs}

\usepackage{hyperref}
\usepackage{url}

\title{JARVIS: From Visual Manuals to Autonomous Robotic Assembly}

\author{
Fan Ding$^{1,2,*}$ \quad
Fucai Ke$^{2,*}$ \quad
Kevin Yuchen Ma$^{1,3}$ \quad
Xuewen Luo$^{4}$ \\
\textbf{
Wenyu Liang$^{1}$ \quad
Vishnu Monn Baskaran$^{2}$ \quad
Hamid Rezatofighi$^{2,\dagger}$ \quad
Yan Wu$^{1,\dagger}$
} \\[0.5em]
$^{1}$A*STAR Institute of Advanced Intelligence and Computing \\
$^{2}$Monash University \quad
$^{3}$National University of Singapore \quad
$^{4}$University of Utah \\
$^{*}$Equal contribution. \quad
$^{\dagger}$Equal contribution.
}

\begin{document}
\setcitestyle{numbers,square,comma}

\maketitle

\begin{abstract}


Visual manuals communicate procedural knowledge through stepwise diagrams, but translating these instructions into reliable long-horizon robotic execution remains challenging. We introduce \ours, a method-agnostic benchmark for autonomous visual-manual-guided execution without additional language instructions. \ours comprises $236{,}679$ LEGO assembly tasks across $8{,}987$ scenes, with a carefully curated \ours-v1 suite of $50$ tasks for interactive robotic evaluation. Tasks span varying initial structural complexities and assembly horizons, requiring agents to interpret manuals, ground actions in an evolving workspace, and execute the intended procedure. A structure-aware evaluation protocol captures geometric accuracy, structural relations, construction order, partial progress, and completion. We further introduce \ours-agentic, a tool-grounded embodied agent that composes robot primitives into executable programs and uses execution feedback for verification and recovery, without task-specific action trajectories or model parameter updates. Across nine multimodal and reasoning models, task-oriented primitives increase the average assembly score. Performance nevertheless declines sharply with assembly horizon, and execution errors remain difficult to correct. Prior-task experience improves initial performance, while repeated reflection provides inconsistent benefits, highlighting persistent limitations in connecting visual procedural reasoning with reliable physical execution.

\end{abstract}

\section{Introduction}
\label{sec:intro}

Across many everyday activities, people learn how to build, assemble, and operate things by following visual manuals. These manuals organize tasks into standardized step-by-step diagrams, providing a compact and transferable representation of procedural knowledge that conveys both the desired outcome and the process required to achieve it~\citep{tie2025manual2skill}. Following such manuals requires humans to interpret the visual instructions, align each illustrated step with the current physical state, and translate the inferred procedure into concrete actions. This ability goes beyond visual recognition alone, requiring visual reasoning to be tightly coupled with sequential decision-making and physical execution~\citep{Machine_Executable_Plan_jiajun}. As multimodal models increasingly move beyond passive perception toward embodied interaction~\citep{ke2026view2space, hong2026esi,ke2025explain}, a fundamental question is whether machines can interpret procedural knowledge from visual manuals and autonomously translate it into action.

Recent studies have taken initial steps toward this capability, particularly in manual-guided robotic assembly~\citep{tie2025manual2skill, Manual_PA_cvpr25, Gu_2026_CVPR_manual_toaction,ke2024hydra}. Existing approaches have primarily developed task-specific solutions for extracting assembly structures, predicting part poses, and generating assembly plans~\citep{tie2025manual2skill, pun2025builable_brick, ke2025dwim, Machine_Executable_Plan_jiajun, yan2025neural_assembler, Li_2026_CVPR_AssemblyBench}. 
Many therefore stop before physical execution; when execution is included, it is often realised through predefined or manually engineered control pipelines. As a result, they do not directly evaluate whether a model can autonomously close the loop from visual manual understanding to physical construction. Benchmarking this capability further requires end-to-end execution tasks and metrics that capture partial progress and structural correctness beyond binary task success~\citep{heo2025furniturebench, wang2025roboeval}. Consequently, it remains unclear how well current multimodal models can use a visual manual as the task specification and autonomously carry out a complete procedure over multiple steps.

\begin{figure*}[t]
    \centering
    \includegraphics[width=\textwidth]{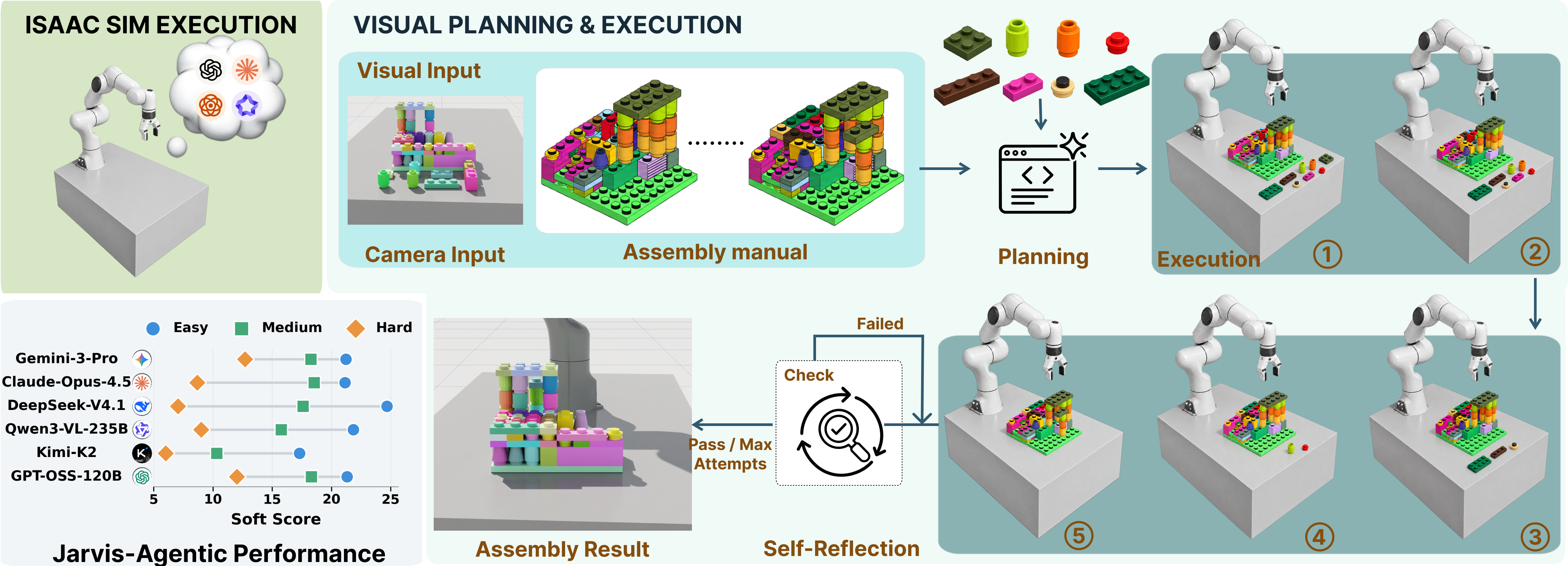}
    \vspace{-1.5em}
    \caption{Overview of the visual planning, execution, and reflection workflow in \ours.}
    \vspace{-1em}
    \label{fig:workflow}
\end{figure*}

To systematically evaluate this capability, we introduce \ours, a method-agnostic benchmark for autonomous visual-manual-guided execution. We construct a large-scale task collection comprising $236{,}679$ assembly tasks across $8{,}987$ LEGO scenes, spanning diverse initial structures and assembly horizons. From this collection, we curate \ours-v1, a suite of $50$ tasks across three difficulty levels for interactive robotic evaluation. As shown in Fig.~\ref{fig:workflow}, an agent receives a visual assembly manual and workspace observations, infers the intended procedure, and executes it while adapting to the evolving physical state. Each manipulation affects subsequent steps, requiring coherent planning and execution throughout assembly. We further introduce a structure-aware evaluation protocol that measures relative geometry, structural relations, and completion, capturing partial progress and accumulated errors beyond binary task success.


Turning manual understanding into physical execution poses a separate practical challenge. Learned control policies typically require task-specific robot action supervision, as low-level trajectories are tied to the robot's kinematics and control interface. Collecting dedicated demonstrations or trajectories for each new construction task is therefore costly. We introduce \ours-agentic, a tool-grounded embodied agent that bridges manual-derived reasoning and physical execution without task-specific action trajectories. Equipped with a compact library of verified robot primitives, the agent grounds the manual, composes executable behaviours, and uses execution reflection for verification, recovery, and replanning.

Using \ours, we evaluate 9 multimodal and reasoning models on manual-guided assembly tasks spanning different structural complexities and assembly horizons. Task-oriented robot skills raise the average assembly score from $7.88$ to $14.71$ without task-specific action trajectories, demonstrating the value of tool-grounded execution. Nevertheless, performance declines sharply with assembly horizon, and improvements in intermediate grounding do not reliably translate into successful assembly. Prior-task experience improves first-round performance, whereas repeated reflection and recovery from execution errors remain unreliable. These findings expose the difficulty of connecting procedural reasoning, physical execution, and reflection-based correction in long-horizon assembly.

Our main contributions are summarized as follows:
\begin{itemize}
    
    \item We formulate \textbf{autonomous visual-manual-guided execution} for long-horizon robotic assembly and introduce \ours, comprising $236{,}679$ tasks across $8{,}987$ scenes, with a curated \ours-v1 suite for interactive evaluation. A structure-aware protocol measures geometric and relational correctness, partial progress, and completion.

    \item We introduce \textbf{\ours-agentic}, a tool-grounded embodied agent that translates visual assembly instructions into executable programs through reusable robot primitives. The framework integrates planning, execution verification, and recovery without task-specific action trajectories or model parameter updates.

    
    \item We evaluate \textbf{nine multimodal and reasoning models}, revealing persistent difficulties in translating manual understanding into reliable long-horizon execution. Our analysis shows that task-oriented tools and prior-task experience improve performance, while online reflection provides inconsistent benefits and error recovery remains limited.
\end{itemize}

\section{Related work}
\label{sec:relatedwork}

\paragraph{Visual assembly planning.}
Visual assembly planning focuses on recovering structural and procedural representations from images. MEPNet predicts stepwise component poses from LEGO manuals~\citep{Machine_Executable_Plan_jiajun}, while Manual-PA estimates assembly order and part poses from instruction diagrams~\citep{Manual_PA_cvpr25}. Neural Assembler infers component graphs and assembly sequences from multiple views of a completed structure~\citep{yan2025neuralAssembleraaai25}, and AssemLM predicts assembly poses from manuals, point clouds, and language~\citep{jing2026assemlm}. Related generative methods produce buildable structures from text~\citep{brickgpt_iccv25_best} or represent brick sequences through connectivity~\citep{Kulits_2026_CVPRBrickNet}. These tasks assess whether a model can infer a valid assembly representation. Autonomous execution additionally requires grounding that representation in a changing workspace and maintaining its validity through physical interaction.

\paragraph{Robotic execution from assembly instructions.}
Several systems connect assembly representations to robot control. Manual2Skill uses assembly graphs, learned pose estimation, and motion planning for furniture assembly~\citep{tie2025manual2skill}, and Manual2Skill++ incorporates explicit connector constraints~\citep{tie2025manual2skillpp}. BrickCraft derives assembly sequences and situated manuals from digital designs to guide learned visuomotor skills~\citep{yu2026brickcraft}. ManualVLA generates intermediate manuals from goal states to condition action execution~\citep{Gu_2026_manualvla_cvpr26}. These systems demonstrate assembly under their respective representations and control designs. A distinct evaluation question is how reliably general-purpose multimodal agents can interpret a supplied visual manual, determine executable actions, and respond to deviations during construction. \ours formulates this capability as a method-agnostic benchmark.

\paragraph{Execution benchmarks and feedback.}
AssemblyBench evaluates assembly order and physically plausible part trajectories~\citep{Li_2026_CVPR_AssemblyBench}, while WorkBenchMark evaluates robotic construction of target LEGO Duplo structures~\citep{ma2026workbenchmark}. CaP-X studies coding agents under different tool abstractions and feedback settings~\citep{fu2026capx}, and ENPIRE investigates policy improvement through repeated physical trials~\citep{xiao2026enpire}. The central question in \ours is whether visual procedural understanding survives execution: can an agent preserve the intended structure as actions accumulate, and recover when the observed state departs from the manual? Our structure-aware evaluation measures partial assembly quality, while controlled experiments examine the effects of horizon, initial complexity, and execution feedback.

\section{\ours Benchmark}
\label{sec:benchmark}

\ours evaluates autonomous visual-manual-guided execution across the full planning–execution pipeline. We instantiate the benchmark with long-horizon LEGO construction in an interactive assembly environment, using visual manuals and structured target states to define each task. This section presents the task formulation, benchmark construction, environment, task composition, and evaluation protocol.

\paragraph{Task Formulation.}
\label{sec:task_formulation}

Let $\mathcal{M}=\{I_0,\ldots,I_T\}$ denote a visual assembly manual, where each image $I_t$ depicts the target construction state $s_t$ at step $t$. The ordered image sequence specifies the desired progression from the initial state $s_0$ to the completed structure $s_T$.
At execution step $t$, the agent receives an observation $o_t$ of the current environment and produces an action $a_t$, inducing the state transition $s_{t+1} = \mathcal{T}(s_t,a_t)$,
where $\mathcal{T}$ denotes the environment dynamics.
The objective is to control the robotic manipulator to realize the final assembly $s_T$, using only the visual manual $\mathcal{M}$ and observations.



\paragraph{Task Composition and Difficulty.}
\label{sec:task_composition}
\ours is constructed from $8,987$ unique LEGO assembly scenes paired with visual manuals and structured 3D models. Rather than treating each complete construction as a single task, we decompose each assembly trajectory into consecutive segments of varying lengths. Each segment defines an independent execution task, consisting of an initial partially assembled state, the corresponding visual manual segment, the required LEGO components, and a target construction state.

This decomposition naturally produces tasks with different execution horizons and scene complexities. In total, \ours contains $236,679$ assembly tasks, each requiring the manipulation of $1$-$43$ LEGO bricks. We further render the same manual instructions in multiple visual styles to evaluate robustness to variations in manual appearance. Details of the data collection, manual-model alignment, and task generation pipeline are provided in the supplementary material.

To characterize task difficulty, we consider two complementary factors: the complexity of the initial partial assembly and the required assembly workload. For task $i$, we define
\begin{equation}
S_i^{\mathrm{scene}}
=
\frac{N_i^{\mathrm{scene}}-N_{\min}^{\mathrm{scene}}}
     {N_{\max}^{\mathrm{scene}}-N_{\min}^{\mathrm{scene}}},
\qquad
S_i^{\mathrm{act}}
=
\frac{N_i^{\mathrm{act}}-N_{\min}^{\mathrm{act}}}
     {N_{\max}^{\mathrm{act}}-N_{\min}^{\mathrm{act}}},
\end{equation}
where $N_i^{\mathrm{scene}}$ is the number of bricks already present in the
pre-assembled structure at the beginning of task $i$, and
$N_i^{\mathrm{act}}$ is the number of additional bricks that the agent must
manipulate and assemble to complete the target structure. The minimum and
maximum values are computed over all tasks in the evaluation benchmark, such
that both normalized measures lie in $[0,1]$.

The overall task difficulty is defined as
\begin{equation}
D_i
=
\alpha S_i^{\mathrm{scene}}
+
(1-\alpha)S_i^{\mathrm{act}},
\label{eq:difficulty_back}
\end{equation}
where we set $\alpha=0.3$, assigning a larger weight to the execution
horizon because manipulation errors can accumulate over successive assembly
steps. Here, $S_i^{\mathrm{scene}}$ measures the normalized total number of
bricks present in the task scene, whereas $S_i^{\mathrm{act}}$ measures the
normalized number of bricks that must be manipulated and assembled.

\begin{table}[t]
\centering
\caption{Statistics of the \ours-v1 benchmark.}
\label{tab:benchmark_statistics}
\resizebox{0.8\linewidth}{!}{%
\begin{tabular}{lccccccc}
\toprule
\multirow{2}{*}{Split}  & \multirow{2}{*}{\#Tasks}
       & \multicolumn{2}{c}{Assembly Horizon} 
       & \multicolumn{2}{c}{Initial Complexity $N_{\mathrm{init}}$}
       & \multicolumn{2}{c}{Total Bricks} \\
\cline{3-4}\cline{5-6}\cline{7-8}
& & avg.& range & avg.& range & avg.& range\\
\midrule
Easy   & 20 & 2.45 & 1-4  & 33.05 & 2-105  & 35.50 & 4-107   \\
Medium & 15 & 4.13  & 2-7  & 78.80  & 2-213  & 82.93  & 7-215  \\
Hard   & 15 & 9.87 & 3-29 & 127.00 & 57-236 & 136.87  & 75-239 \\
\bottomrule
\end{tabular}%
}
\end{table}

From the full task pool, we first curate a representative candidate set of $1,067$ validation tasks, selected to cover diverse source assemblies, construction scales, and execution horizons. We then rank these candidates by $D_i$ and partitioned into three difficulty strata: the bottom 40\% constitute the Easy stratum, the next 30\% the Medium stratum, and the top 30\% the Hard stratum. Within each stratum, we sample tasks at approximately uniform intervals along the difficulty distribution while promoting diversity among the source LEGO assemblies. The resulting \ours-v1 comprises 50 interactive tasks: 20 Easy, 15 Medium, and 15 Hard as shown in Table\ref{tab:benchmark_statistics}.

\paragraph{Interactive Assembly Environment.}
\label{sec:simulation}

Each task is instantiated as an interactive robotic assembly episode in NVIDIA Isaac Sim 6.0.1~\citep{pmlr_isaac_sim}, using a Franka Emika Panda manipulator~\citep{gaz2019franka_emika_panda} with a parallel-jaw gripper. Each scene contains a partially assembled LEGO structure and the movable bricks required by the corresponding manual segment. The agent receives top-down RGB observations and robot-state feedback, and interacts with the arm and gripper through ROS~2~\citep{macenski2022ros_2_robot}. Robot motion is executed with NVIDIA cuMotion/RMPflow, with MoveIt~\citep{chitta2016moveit} supported as an alternative backend.

During each episode, the evaluated method receives the visual manual and current scene observations and must autonomously complete the assembly until termination or the execution budget is exhausted. Ground-truth structure, object poses, and target placements are reserved for evaluation and are not exposed to the method, except for explicitly defined oracle baselines. Further implementation details are provided in the supplementary material.

\paragraph{Structure-Aware Assembly Evaluation.}
\label{sec:evaluation}

A binary success criterion cannot fully capture assembly quality. We therefore adopt a structure-aware evaluation that separately measures geometric accuracy, task completion, and final-state validity. Specifically, the pose score $S_{\mathrm{pose}}$ evaluates each manipulated brick with respect to its intended support, accounting for relative position, vertical alignment, symmetry-aware orientation, and a low-weight absolute-position anchor. Completion $S^{\mathrm{comp}}$ captures how much of the required assembly has been achieved.
These components are combined into an overall soft score, with a bounded time penalty applied only when execution exceeds a task-dependent budget.

\textit{Pose accuracy.}
We evaluate each required brick against its target pose in the simulator world frame. For brick $j$, we compute its Euclidean position error $e_j^{\mathrm{pos}}$, vertical error $e_j^z$, and angular orientation error $e_j^\theta$. Each error is converted into a soft score using a Gaussian kernel,
\begin{equation}
    s_j^{k}
    =
    \exp\left[
        -\frac{1}{2}
        \left(\frac{e_j^{k}}{\sigma_k}\right)^2
    \right],
    \qquad
    k\in\{\mathrm{pos},z,\theta\},
\end{equation}
where $\sigma_k$ controls the tolerance of each term. Missing bricks and bricks that have not been moved from their initial poses receive zero credit. We average the component scores over the $N_i$ required bricks in task $i$:
\begin{equation}
    S_i^{\mathrm{pose}}
    =
        \frac{1}{N_i}
    \sum_{j=1}^{N_i}
    \left(
    \lambda_{\mathrm{pos}}s_j^{\mathrm{pos}}
    +
    \lambda_z s_j^z
    +
    \lambda_{\mathrm{\theta}}s_j^{\mathrm{\theta}}\right),
\end{equation}
where
$\lambda_{\mathrm{pos}}+\lambda_z+\lambda_{\mathrm{\theta}}=1$.
This formulation provides partial credit for near-correct placements while smoothly penalising larger errors. The kernel bandwidths and component weights are reported in the supplementary material.

\textit{Completion.}
The completion score measures the fraction of required bricks successfully placed near their target positions. A brick is counted as completed if it has been moved by more than $5\,\mathrm{mm}$ from its initial pose, its Euclidean distance to the target position does not exceed $5\,\mathrm{cm}$, and it is not held by the gripper when the episode terminates. Thus,
\begin{equation}
    S_i^{\mathrm{comp}}
    =
    \frac{1}{N_i}
    \sum_{j=1}^{N_i}
    \mathbb{I}
    \left[
        d_j^{\mathrm{move}}>5\,\mathrm{mm}
        \,\land\,
        e_j^{\mathrm{pos}}\leq5\,\mathrm{cm}
        \,\land\,
        \neg h_j
    \right],
\end{equation}
where $N_i$ is the number of required bricks in task $i$, $d_j^{\mathrm{move}}$ is the displacement from the initial pose, and $h_j$ indicates whether the brick remains held by the gripper.

\textit{Final-state validity.}
We consider whether each manipulated brick remains in a valid state at the end of the episode. A brick receives validity credit if it has been moved beyond the movement tolerance, its horizontal bounding-box projection remains within the designated tabletop region, and it is no longer held by the gripper.

\textit{Overall score.}
Before accounting for execution time, the assembly-quality score for task $i$ is
\begin{equation}
    S^{\mathrm{quality}}_i
    =
    \lambda_{{pose}}S_i^{\mathrm{pose}}
    +
    \lambda_{\mathrm{comp}}S_i^{\mathrm{comp}}
\end{equation}
where
$\lambda_{\mathrm{pose}}+\lambda_{\mathrm{comp}}=1$.
The score is set to zero if no required brick is moved beyond the movement tolerance. We define a task-dependent time budget as
\begin{equation}
    T_i
    =
    T_{\mathrm{base}}
    +
    N_iT_{\mathrm{brick}},
\end{equation}
where $T_{\mathrm{base}}$ is the base time allowance, and $T_{\mathrm{brick}}$ is the additional allowance per brick. Let $t_i$ denote the agent's actual execution time for task $i$. The final episode score combines assembly quality with a bounded time penalty:
\begin{equation}
S_i
=
100S_i^{\mathrm{quality}}
\left[
\alpha+(1-\alpha)\exp\left(-\frac{\max(0,t_i-T_i)}{T_i}\right)
\right],
\end{equation}
where the time-dependent factor mildly penalizes slower execution while remaining bounded. The corresponding hyperparameters are reported in the supplementary material.

\section{\ours-agentic}
\label{sec:methodology}

\begin{figure*}[t]
    \centering
    \includegraphics[width=\textwidth]{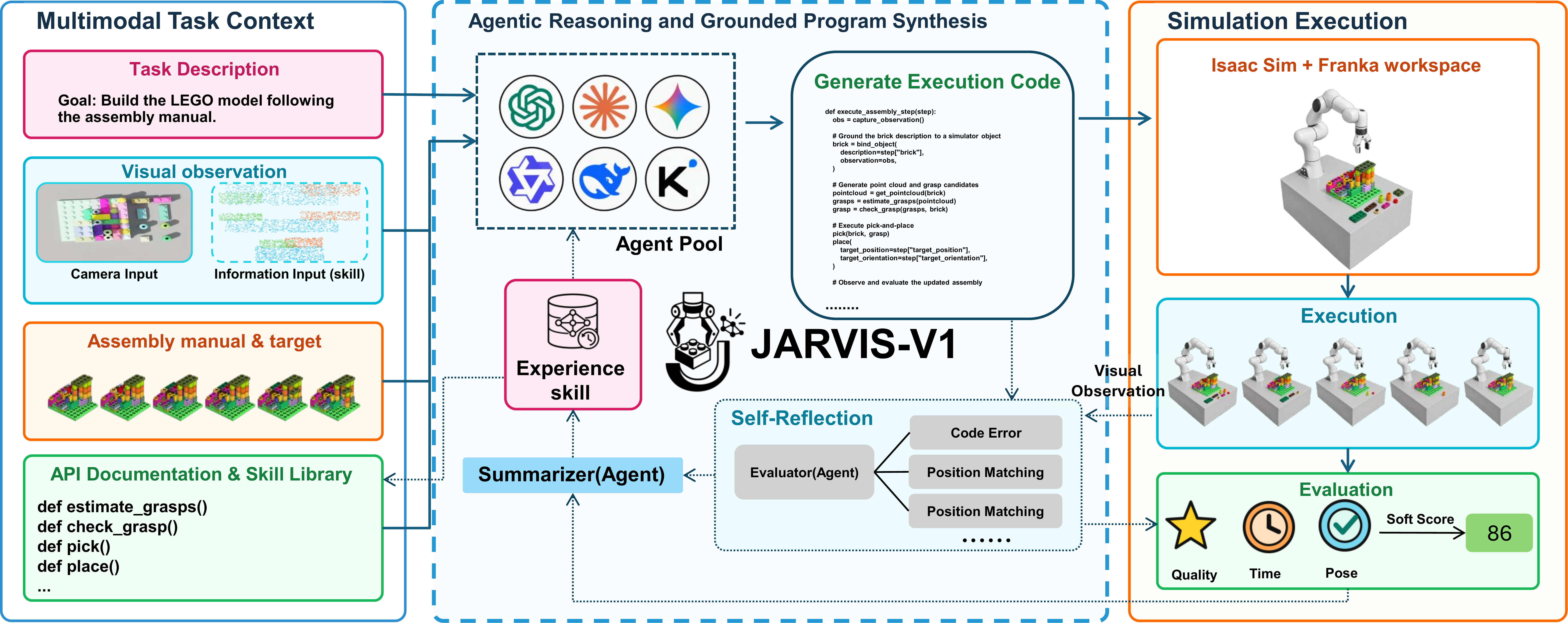}
    \caption{Overview of JARVIS-V1. The framework integrates multimodal task context, agentic code generation, simulation execution, evaluation, and self-reflection in a closed-loop assembly process.}
    \label{fig:jarvis_method}
\end{figure*}


We introduce \ours-agentic, a tool-grounded vision-language-action framework for long-horizon LEGO assembly as shown in Figure~\ref{fig:jarvis_method}. Given a visual assembly manual and the current workspace observation, the agent assesses assembly progress, identifies the next step, and generates executable code using constrained robot APIs and reusable manipulation skills. After execution, it observes the resulting scene and operation feedback, allowing it to verify progress, recover from failures, and replan over multiple interaction rounds.

The framework integrates planning and execution while decoupling the agent's decisions from low-level robot control. Adapting VLA policies to new robot setups and manipulation tasks often requires demonstrations and fine-tuning~\citep{kim2024openvla}, making trajectory collection costly for manual-guided assembly. In \ours-agentic, the agent interprets the manual, grounds intended actions in the current scene, and composes skills into executable programs; the skill library handles grasping, constrained motion, placement, and operation-level checks. This division lets an LLM control the full assembly process through interaction with the environment, without predicting robot trajectories or training a task-specific action policy.



\paragraph{Grounded Program Generation.}

The adapted agent turns a manual-derived assembly step into an executable sequence of tool calls. It queries the perception APIs to locate the relevant bricks and assess the current assembly state, then grounds the intended manipulation in the observed workspace. Based on these results, it generates Python code that selects and parameterizes skills for grasping, transporting, and placing the brick. The skills execute the underlying robot motions and return observable outcomes, which the agent can use to verify the operation and determine its next step. The agent composes existing robot capabilities through code, enabling assembly without training a task-specific action policy or collecting robot action trajectories.

Robot motions are carried out through the motion-planning backend in Isaac Sim~\cite{pmlr_isaac_sim}. API definitions, skill implementations, and program constraints are provided in the supplementary material.

\paragraph{Execution Memory and Verification.}

After each execution round, the agent receives updated observations and feedback on the attempted operation. A structured memory records the generated plan and code, API calls, observed object states, and relevant grasp, placement, and runtime outcomes. Before the next reasoning step, this history is compressed into a task-relevant summary. The agent can then assess whether the intended change occurred and, if necessary, revise its next action using the current scene rather than its previous plan alone. Ground-truth target poses, target-scene states, and final evaluation scores remain accessible only to the external evaluator.

\paragraph{Reflection from Execution Experience.}

We consider two optional reflection settings. In \emph{Standard Reflection}, the agent uses observations, generated code, and execution feedback from the current task to diagnose an unsuccessful attempt and revise its program. \emph{Reflection + Prior-Task} additionally provides task-independent lessons summarized from execution logs of training tasks, including common failures, effective strategies, and API usage. These lessons inform subsequent decisions while the agent continues to reflect on feedback from the current task. The comparison isolates the benefit of transferring execution experience across tasks. Neither setting updates model parameters. Memory construction and reflection prompts are detailed in the supplementary material.

\section{Experimental Results \& Analysis}
\label{sec:exp_res}

We evaluate how contemporary multimodal and reasoning models perform when embedded in the \ours-agentic framework for full pipeline visual-manual-guided assembly. Our experiments examine not only overall task performance, but also the factors that determine successful physical execution. Particularly, we study whether strong visual planning translates into reliable execution, how agents respond to execution errors, and whether additional reflection improves long-horizon assembly.



\paragraph{Effect of Reasoning Models.}
\label{sec:vlm_tools}
We first benchmark $9$ contemporary multimodal and reasoning models using the same \ours-agentic execution framework and interaction budget. Table~\ref{tab:model_tool_comparison} reports their assembly performance across tasks of varying construction complexity and assembly horizon.

Tool abstraction has a substantial impact on execution performance. Replacing generic robot-control APIs with our task-oriented primitives improves the Overall score for all nine models, raising the average from $7.88$ to $14.71$. The gain is largest on Medium tasks, where the average score increases from $6.29$ to $15.17$. On Hard tasks, the improvement is more modest ($4.49 \rightarrow 8.35$), suggesting that better execution abstractions alone are insufficient once long-horizon reasoning and state tracking become dominant bottlenecks.

The tool interface also changes the relative model ranking. Claude-Opus-4.5~\citep{anthropic2026claudeopus} leads with generic APIs, whereas DeepSeek-V4.1~\citep{xu2026deepseek41flash} achieves the highest Overall score ($17.39$) with task-oriented primitives. Notably, the best Overall performance does not coincide with the fewest failed actions: GPT-4o~\citep{hurst2024gpt4o} makes fewer failed actions per task than DeepSeek-V4.1, yet attains a lower Overall score. This suggests that assembly performance depends not only on avoiding invalid operations, but also on whether successful actions produce meaningful structural progress and whether the agent can recover from intermediate mistakes. More broadly, evaluations under low-level APIs may partly conflate reasoning ability with tool orchestration.


\begin{table*}[t]
\centering
\caption{Model and tool-library performance across task difficulty levels. \underline{Underline} and \textcolor{blue}{blue} denote the best generic and task-oriented results; \textbf{bold} marks the best Overall score within each library.}
\label{tab:model_tool_comparison}
{\renewcommand{\arraystretch}{0.85}%
\resizebox{\textwidth}{!}{%
\begin{tabular}{ll|ccc|c|c|c}
\toprule
\multirow{2}{*}{Model} & \multirow{2}{*}{Tool library} & \multicolumn{4}{c}{Soft Score $\uparrow$} & \multirow{2}{*}{Valid (\%) $\uparrow$} & \multirow{2}{*}{\makecell{Failed actions\\per task $\downarrow$}} \\
\cmidrule(lr){3-6}
& & \textbf{Easy} & \textbf{Medium} & \textbf{Hard} & \textbf{Overall} & & \\
\midrule
\rowcolor{rowshade}
\cellcolor{white}\multirow{2}{*}{Gemini-3-Pro~\citep{Google2026Gemini3}} & Generic & 13.27 & 6.61 & 4.68 & 8.19 & 98.00 & 6.04 \\
& Task-oriented & 22.16 & \textcolor{blue}{19.30} & 9.94 & 17.13 & 100.00 & 4.96 \\
\midrule
\rowcolor{rowshade}
\cellcolor{white}\multirow{2}{*}{Claude-Opus-4.5~\citep{anthropic2026claudeopus}} & Generic & \underline{15.84} & \underline{16.13} & \underline{9.51} & \underline{\textbf{13.83}} & 92.00 & 7.36 \\
& Task-oriented & 22.20 & 18.58 & \textcolor{blue}{10.35} & 17.04 & 100.00 & 5.31 \\
\midrule
\rowcolor{rowshade}
\cellcolor{white}\multirow{2}{*}{Claude-Haiku-4.5~\citep{Anthropic2026Haiku45}} & Generic & 11.69 & 3.22 & 2.88 & 5.93 & 90.00 & 6.70 \\
& Task-oriented & 19.51 & 9.41 & 6.12 & 11.68 & 96.00 & 6.10 \\
\midrule
\rowcolor{rowshade}
\cellcolor{white}\multirow{2}{*}{DeepSeek-V4.1~\citep{xu2026deepseek41flash}} & Generic & 15.37 & 6.33 & 3.76 & 8.49 & 90.00 & 6.50 \\
& Task-oriented & \textcolor{blue}{25.66} & 18.52 & 7.98 & \textcolor{blue}{\textbf{17.39}} & 96.00 & 4.26 \\
\midrule
\rowcolor{rowshade}
\cellcolor{white}\multirow{2}{*}{Qwen3-VL-235B~\citep{bai2025qwen3vl}} & Generic & 13.78 & 6.01 & 5.03 & 8.27 & 90.00 & 6.50 \\
& Task-oriented & 21.85 & 15.75 & 9.01 & 15.54 & 94.00 & 5.30 \\
\midrule
\rowcolor{rowshade}
\cellcolor{white}\multirow{2}{*}{Kimi-K2.5~\citep{team2026kimik25}} & Generic & 10.37 & 3.53 & 2.83 & 5.58 & 88.00 & 6.90 \\
& Task-oriented & 17.31 & 10.31 & 6.01 & 11.21 & 94.00 & 5.90 \\
\midrule
\rowcolor{rowshade}
\cellcolor{white}\multirow{2}{*}{GPT-4o~\citep{hurst2024gpt4o}} & Generic & 10.29 & 2.64 & 0.93 & 4.62 & 24.00 & 0.72 \\
& Task-oriented & 16.21 & 11.23 & 7.21 & 11.55 & 70.00 & 3.50 \\
\midrule
\rowcolor{rowshade}
\cellcolor{white}\multirow{2}{*}{GPT-5.1~\citep{openai2026gpt51}} & Generic & 12.38 & 5.88 & 4.64 & 7.63 & 90.00 & 6.80 \\
& Task-oriented & 20.04 & 16.26 & 8.66 & 14.99 & 96.00 & 14.84 \\
\midrule
\cellcolor{white}\multirow{2}{*}{GPT-OSS-120B$^{*}$~\citep{agarwal2025gptoss}} & Generic & 12.77 & 6.26 & 6.12 & 8.38 & 92.00 & 6.50 \\
& Task-oriented & 20.66 & 17.20 & 9.85 & 15.90 & 96.00 & 5.20 \\
\bottomrule
\end{tabular}%
}
}
\vspace{1mm}
\parbox{\textwidth}{\scriptsize $^{*}$For GPT-OSS-120B, instruction-manual understanding is performed by GPT-5.1, while subsequent planning and execution are handled by GPT-OSS-120B. Valid (\%) denotes the proportion of generated programs that can be executed. Overall is the unweighted mean across the three difficulty levels.}
\end{table*}

\paragraph{What makes long-horizon assembly difficult?}
\label{sec:complexity}

Our study shows that assembly difficulty is jointly determined by assembly horizon and initial structural complexity, with current models appearing particularly sensitive to increases in horizon. As shown in Table~\ref{tab:complexity_horizon}, under low initial complexity, increasing the assembly horizon from $1$-$2$ to $3$-$5$ bricks reduces the macro-average score sharply from $40.01$ to $18.29$.. Similar degradation is observed across other complexity regimes, suggesting that even a moderate increase in sequential manipulation steps can substantially increase error accumulation and planning difficulty. Initial structure complexity also affects performance: for short-horizon tasks, the macro-average score decreases from $40.01$ to $16.64$ as the initial structure grows from $0$-$29$ to $79+$ bricks. However, this effect is generally more gradual and less consistent across individual bins, indicating a stronger dependence on the particular structure and spatial configuration being manipulated.
Performance is lowest when both factors are high, suggesting that errors in grounding and execution become harder to contain over longer assembly sequences. The task-distribution heatmap in the supplementary material shows that the Easy, Medium, and Hard partitions span this joint difficulty gradient. Across complexity regimes, assembly horizon remains the more consistent source of performance degradation.

\begin{table*}[t]
\centering
\caption{Joint effect of initial scene complexity and assembly horizon.
Each cell reports the mean soft score, with the number of evaluated tasks in
parentheses. Results are computed from the latest no-reflection execution
round for each model.}
\label{tab:complexity_horizon}

{\renewcommand{\arraystretch}{0.85}%
\resizebox{0.7\textwidth}{!}{%
\begin{tabular}{llccc}
\toprule
& &
\multicolumn{3}{c}{Assembly horizon $N_{\mathrm{act}}$} \\
\cmidrule(lr){3-5}
Model
& Initial complexity $N_{\mathrm{init}}$
& 1--2 & 3--5 & 6+ \\
\midrule

\multirow{3}{*}{Gemini-3-Pro~\citep{Google2026Gemini3}}
& 0--29  & 43.42 (5) & 15.40 (10) & -- \\
& 30--78 & 28.03 (4) & 17.99 (8)  & 10.40 (6) \\
& 79+    & 16.40 (4) & 12.96 (7)  & 6.01 (6) \\
\midrule

\multirow{3}{*}{Claude-Opus-4.5~\citep{anthropic2026claudeopus}}
& 0--29  & 36.46 (5) & 18.38 (10) & -- \\
& 30--78 & 30.87 (4) & 14.47 (8)  & 11.88 (6) \\
& 79+    & 17.61 (4) & 14.05 (7)  & 5.44 (6) \\
\midrule

\multirow{3}{*}{DeepSeek-V4.1~\citep{xu2026deepseek41flash}}
& 0--29  & 40.16 (5) & 21.08 (10) & -- \\
& 30--78 & 31.25 (4) & 19.47 (8)  & 9.78 (6) \\
& 79+    & 15.91 (4) & 12.12 (7)  & 1.83 (6) \\
\midrule

\multirow{3}{*}{\textbf{Macro average}}
& 0--29  & \textbf{40.01} & \textbf{18.29} & -- \\
& 30--78 & \textbf{30.05} & \textbf{17.31} & \textbf{10.69} \\
& 79+    & \textbf{16.64} & \textbf{13.05} & \textbf{4.43} \\
\bottomrule
\end{tabular}%
}
}
\end{table*}

\begin{table*}[t]
\centering
\caption{Manual planning and spatial grounding before execution. Matched and missing steps count correctly identified and omitted assembly steps; coverage is the fraction of required steps matched. Position and orientation errors measure placement accuracy over matched steps. Position and orientation errors are computed on matched steps. 3D denotes structured world-frame descriptors added to RGB-D observations. More details about 3D descriptors are in supplementary.}
\label{tab:benchmark50_grounding_comparison}

{\renewcommand{\arraystretch}{0.9}%
\resizebox{1\textwidth}{!}{%
\begin{tabular}{llccccc}
\toprule
Model
& Input
& \makecell{Position Error\\(mm) $\downarrow$}
& \makecell{Orientation Error\\($^\circ$) $\downarrow$}
& \makecell{Matched\\Steps $\uparrow$}
& \makecell{Missing\\Steps $\downarrow$}
& \makecell{Coverage\\(\%) $\uparrow$} \\
\midrule

\multirow{2}{*}{Gemini-3-Pro~\citep{Google2026Gemini3}}
& RGB-D observation
& 78.08 & 16.87 & \textbf{40} & \textbf{219} & \textbf{15.44} \\
& RGB-D + 3D descriptors
& \textbf{16.76} & \textbf{12.00} & 34 & 225 & 13.13 \\

\midrule

\multirow{2}{*}{Claude-Opus-4.5~\citep{anthropic2026claudeopus}}
& RGB-D observation
& 49.49 & \textbf{18.00} & \textbf{65} & \textbf{194} & \textbf{25.10} \\
& RGB-D + 3D descriptors
& \textbf{29.09} & 23.57 & 63 & 196 & 24.32 \\

\midrule

\multirow{2}{*}{Claude-Haiku-4.5~\citep{Anthropic2026Haiku45}}
& RGB-D observation
& 184.12 & 20.96 & \textbf{50} & \textbf{209} & \textbf{19.31} \\
& RGB-D + 3D descriptors
& \textbf{30.34} & \textbf{16.67} & 29 & 230 & 11.20 \\

\midrule

\multirow{2}{*}{DeepSeek V4.1~\cite{xu2026deepseek41flash}}
& RGB-D observation
& 179.84
& \textbf{16.36}
& 11
& 248
& 4.25 \\

& RGB-D + 3D descriptors
& \textbf{23.00}
& 18.97
& \textbf{26}
& \textbf{233}
& \textbf{10.04} \\

\midrule

\multirow{2}{*}{GPT-4o~\citep{hurst2024gpt4o}}
& RGB-D observation
& 189.64 & \textbf{9.57} & 47 & 212 & 18.15 \\
& RGB-D + 3D descriptors
& \textbf{22.39} & 13.49 & \textbf{48} & \textbf{211} & \textbf{18.53} \\

\midrule

\multirow{2}{*}{GPT-5.1~\citep{openai2026gpt51}}
& RGB-D observation
& 154.09 & \textbf{22.50} & 16 & 243 & 6.18 \\
& RGB-D + 3D descriptors
& \textbf{20.36} & 41.57 & \textbf{33} & \textbf{226} & \textbf{12.74} \\

\midrule

\multirow{2}{*}{Qwen3-VL-235B~\citep{bai2025qwen3vl}}
& RGB-D observation
& 171.39 & \textbf{5.30} & 17 & 242 & 6.56 \\
& RGB-D + 3D descriptors
& \textbf{21.42} & 9.68 & \textbf{32} & \textbf{227} & \textbf{12.36} \\

\midrule

\multirow{2}{*}{Kimi-K2.5~\citep{team2026kimik25}}
& RGB-D observation
& 178.73 & \textbf{18.46} & 26 & 233 & 10.04 \\
& RGB-D + 3D descriptors
& \textbf{21.67} & 23.29 & \textbf{41} & \textbf{218} & \textbf{15.83} \\

\bottomrule
\end{tabular}%
}
}
\end{table*}

\paragraph{Why Full-Pipeline Evaluation Matters.}
\label{sec:full_pipeline}
Planning and grounding scores alone do not capture whether an agent can complete an assembly. Table~\ref{tab:benchmark50_grounding_comparison} shows that models can achieve relatively accurate placement estimates for matched steps, while the best model covers only $25.10\%$ of the required steps. Structured 3D descriptors further reduce position error for several models (e.g., $78.08$ to $16.76$\,mm for Gemini-3-Pro), but these intermediate improvements do not ensure successful execution. In the complete pipeline, the best Overall assembly score is only $17.39$ (Table~\ref{tab:model_tool_comparison}). Manipulation and placement failures, state changes, and error accumulation introduce challenges that planning metrics cannot measure. We therefore evaluate the assembled outcome, with execution-stage analysis provided in the supplementary material.

\paragraph{Can Agents Recover from Execution Errors?}
\label{sec:recovery}
Across the $25$ perturbed tasks, the average score increases only from $54.73$ to $55.14$ ($+0.42$ points), indicating limited overall recovery. Outcomes vary substantially across tasks: only $8$ improve, although the largest gain reaches $26.22$ points, while the largest decline is $15.99$ points. Manual inspection suggests that successful recovery tends to occur when the assembly has lower structural complexity at the point of failure, consistent with the performance trends observed above. Task-level results are reported in the supplementary material.

\begin{table*}[t]
\centering
\caption{Performance across reflection rounds with the task-oriented tool library. R1-R3 denote scores after one to three reflection rounds, respectively. Prior-task summaries are distilled from completed assembly tasks. Percentage changes are relative to the preceding round, and Overall is the macro-average across Low, Medium, and High. Blue highlights the largest improvement in each percentage column.}
\label{tab:reflection_comparison}
{\renewcommand{\arraystretch}{1}%
\resizebox{\textwidth}{!}{%
\begin{tabular}{lllrrrrrrrrrr}
\toprule
& & &
\multicolumn{5}{c}{w/ Time Penalty} &
\multicolumn{5}{c}{w/o Time Penalty} \\
\cmidrule(lr){4-8}
\cmidrule(lr){9-13}
Model
& Reflection Framework
& Difficulty
& R1
& R2
& R3
& R1$\rightarrow$R2
& R2$\rightarrow$R3
& R1
& R2
& R3
& R1$\rightarrow$R2
& R2$\rightarrow$R3 \\
& & &
& & &
(\%)
& (\%)
& & & &
(\%)
& (\%) \\
\midrule

&
& Easy
& 13.926 & 9.922 & 9.361
& $-28.75$ & $-5.65$
& 14.311 & 13.079 & 11.220
& $-8.61$ & $-14.21$ \\

& & Medium
& 12.617 & 11.352 & 10.241
& $-10.03$ & $-9.79$
& 13.651 & 12.320 & 11.760
& $-9.75$ & $-4.55$ \\

& 
& Hard
& 11.770 & 11.891 & 9.218
& $+1.03$ & $-22.48$
& 12.452 & 13.475 & 10.703
& $+8.22$ & $-20.57$ \\

\rowcolor{rowshade}
\cellcolor{white}
& \cellcolor{white}\multirow{-4}{*}{Standard Reflection}
& \textbf{Overall}
& \textbf{12.771} & \textbf{11.055} & \textbf{9.607}
& $\mathbf{-13.44}$ & $\mathbf{-13.10}$
& \textbf{13.471} & \textbf{12.958} & \textbf{11.228}
& $\mathbf{-3.81}$ & $\mathbf{-13.35}$ \\

\cmidrule(lr){2-13}

& \multirow{4}{*}{%
    \makecell[l]{Reflection + Prior-Task\\Experience Summary}}
& Easy
& 21.370 & 22.450 & 19.450
& $+5.05$ & $-13.36$
& 24.920 & 26.770 & 22.130
& $+7.42$ & $-17.33$ \\

& & Medium
& 20.700 & 21.500 & 20.060
& $+3.86$ & $-6.70$
& 23.320 & 24.520 & 23.050
& $+5.15$ & $-6.00$ \\

& 
& Hard
& 13.820 & 13.200 & 13.200
& $-4.49$ & $0.00$
& 13.910 & 13.800 & 13.800
& $-0.79$ & $0.00$ \\

\rowcolor{rowshade}
\cellcolor{white}\multirow{-8}{*}{Gemini-3-Pro}
&\cellcolor{white}
& \textbf{Overall}
& \textbf{18.630} & \textbf{19.050} & \textbf{17.570}
& $\mathbf{+2.25}$ & $\mathbf{-7.77}$
& \textbf{20.720} & \textbf{21.700} & \textbf{19.660}
& $\mathbf{+4.73}$ & $\mathbf{-9.40}$ \\

\midrule

& 
& Easy
& 14.280 & 18.070 & 18.040
& $+26.54$ & $-0.17$
& 16.510 & 22.510 & 22.500
& $+36.34$ & $-0.04$ \\

& & Medium
& 16.610 & 13.570 & 13.370
& $-18.30$ & $-1.47$
& 19.250 & 16.520 & 16.510
& $-14.18$ & $-0.06$ \\

&
& Hard
& 10.600 & 9.920 & 9.620
& $-6.42$ & $-3.02$
& 10.740 & 10.190 & 9.900
& $-5.12$ & $-2.85$ \\

\rowcolor{rowshade}
\cellcolor{white}
&\cellcolor{white} \multirow{-4}{*}{Standard Reflection}
& \textbf{Overall}
& \textbf{13.830} & \textbf{13.850} & \textbf{13.680}
& $\mathbf{+0.14}$ & $\mathbf{-1.23}$
& \textbf{15.500} & \textbf{16.410} & \textbf{16.310}
& $\mathbf{+5.87}$ & $\mathbf{-0.61}$ \\

\cmidrule(lr){2-13}

&\multirow{4}{*}{%
    \makecell[l]{Reflection + Prior-Task\\Experience Summary}}
& Easy
& 25.070 & 15.890 & 16.000
& $-36.62$ & $+0.69$
& 28.250 & 18.970 & 19.420
& $-32.85$ & $+2.37$ \\

& & Medium
& 19.010 & 17.280 & 17.200
& $-9.10$ & $-0.46$
& 20.780 & 19.850 & 19.880
& $-4.48$ & $+0.15$ \\

&
& Hard
& 10.570 & 10.290 & 10.730
& $-2.65$ & $+4.28$
& 11.000 & 11.130 & 11.770
& $+1.18$ & $+5.75$ \\

\rowcolor{rowshade}
\cellcolor{white}\multirow{-8}{*}{Claude-Opus-4.5}
&\cellcolor{white}
& \textbf{Overall}
& \textbf{18.220} & \textbf{14.490} & \textbf{14.640}
& $\mathbf{-20.47}$ & $\mathbf{+1.04}$
& \textbf{20.010} & \textbf{16.650} & \textbf{17.020}
& $\mathbf{-16.79}$ & \textcolor{blue}{$\mathbf{+2.22}$} \\

\midrule

& 
& Easy
& 7.950 & 11.260 & 10.060
& $+41.64$ & $-10.66$
& 8.450 & 13.860 & 12.640
& $+64.02$ & $-8.80$ \\

& & Medium
& 11.490 & 15.620 & 19.880
& $+35.94$ & $+27.27$
& 15.940 & 21.130 & 21.260
& $+32.56$ & $+0.62$ \\

&
& Hard
& 10.090 & 12.230 & 12.040
& $+21.21$ & $-1.55$
& 10.180 & 13.290 & 13.290
& $+30.55$ & $0.00$ \\

\rowcolor{rowshade}
\cellcolor{white}
&\cellcolor{white} \multirow{-4}{*}{Standard Reflection}
& \textbf{Overall}
& \textbf{9.840} & \textbf{13.040} & \textbf{13.990}
& \textcolor{blue}{$\mathbf{+32.52}$} & \textcolor{blue}{$\mathbf{+7.29}$}
& \textbf{11.520} & \textbf{16.090} & \textbf{15.730}
& \textcolor{blue}{$\mathbf{+39.67}$} & $\mathbf{-2.24}$ \\

\cmidrule(lr){2-13}

&\multirow{4}{*}{%
    \makecell[l]{Reflection + Prior-Task\\Experience Summary}}
& Easy
& 25.820 & 22.690 & 23.410
& $-12.12$ & $+3.17$
& 30.730 & 29.270 & 30.470
& $-4.75$ & $+4.10$ \\

& & Medium
& 15.330 & 17.210 & 18.310
& $+12.26$ & $+6.39$
& 23.310 & 21.660 & 22.310
& $-7.08$ & $+3.00$ \\

&& Hard
& 12.810 & 10.960 & 10.060
& $-14.44$ & $-8.21$
& 13.480 & 12.260 & 11.460
& $-9.05$ & $-6.53$ \\

\rowcolor{rowshade}
\cellcolor{white}\multirow{-8}{*}{DeepSeek-V4.1}
&\cellcolor{white} & \textbf{Overall}
& \textbf{17.990} & \textbf{16.950} & \textbf{17.260}
& $\mathbf{-5.78}$ & $\mathbf{+1.83}$
& \textbf{22.510} & \textbf{21.060} & \textbf{21.410}
& $\mathbf{-6.44}$ & $\mathbf{+1.66}$ \\

\bottomrule
\end{tabular}%
}
}
\end{table*}

\paragraph{How Much Does Closed-Loop Feedback Contribute to Long-Horizon Execution?}
We compare standard reflection and reflection augmented with prior-task experience against the no-reflection results in Table~\ref{tab:reflection_comparison}. Without reflection, Gemini-3-Pro, Claude-Opus-4.5, and DeepSeek-V4.1 score $17.13$, $17.04$, and $17.39$, respectively as shown in Table~\ref{tab:model_tool_comparison}. Under standard reflection, their R1 scores fall to $12.77$, $13.83$, and $9.84$ with the time penalty. Even after removing the penalty, all three R1 scores remain below the no-reflection results. In contrast, prior-task experience raises the time-penalized R1 scores to $18.63$, $18.22$, and $17.99$.

Further reflection rarely closes this gap. DeepSeek-V4.1 improves under standard reflection from $9.84$ at R1 to $13.04$ at R2, but remains below its no-reflection score; additional rounds provide little consistent benefit across models. These results indicate that current agents struggle to turn feedback from spatially and physically grounded execution into reliable corrections. Summarized experience from previous tasks is more useful than repeated reflection on the current attempt.




\section{Conclusion}

We introduced \ours, a benchmark for autonomous visual-manual-guided execution with structure-aware evaluation, and \ours-agentic, a tool-grounded agent requiring no task-specific action trajectories. Experiments show that robot skills and prior-task experience improve assembly performance, while longer horizons, execution errors, and unreliable reflection remain substantial challenges. These findings motivate agents that can more reliably connect visual procedural knowledge with physical execution and error correction.

\subsection*{AI use statement}
In this work, we used generative AI tools to obtain feedback on the conceptual framing, assist with software development and debugging. We also used them for grammar and clarity edits and suggestions on figure colour schemes.

We did not use generative AI to generate synthetic datasets, formulate mathematical claims, construct or prove theoretical results, or clean or reformat datasets. These tasks were either not applicable or were completed without generative AI assistance.

The authors reviewed all AI-assisted suggestions, code, and edits. We take responsibility for the final content of this work.

\subsubsection*{Acknowledgments}
This work was partially supported by the ONR Global X-Challenge Grant (N62909-25-1-2067).

The majority of this work was conducted during Fan Ding's research internship at the A*STAR Institute of Advanced Intelligence and Computing (A*STAR), Singapore.


\bibliography{iclr2027_conference}
\bibliographystyle{unsrtnat}

\appendix
\clearpage
\setcounter{page}{1}

\section*{Supplementary Material}

\section{Simulated Assembly Tasks}

\begin{figure*}[t]
    \centering
    \includegraphics[width=\textwidth]{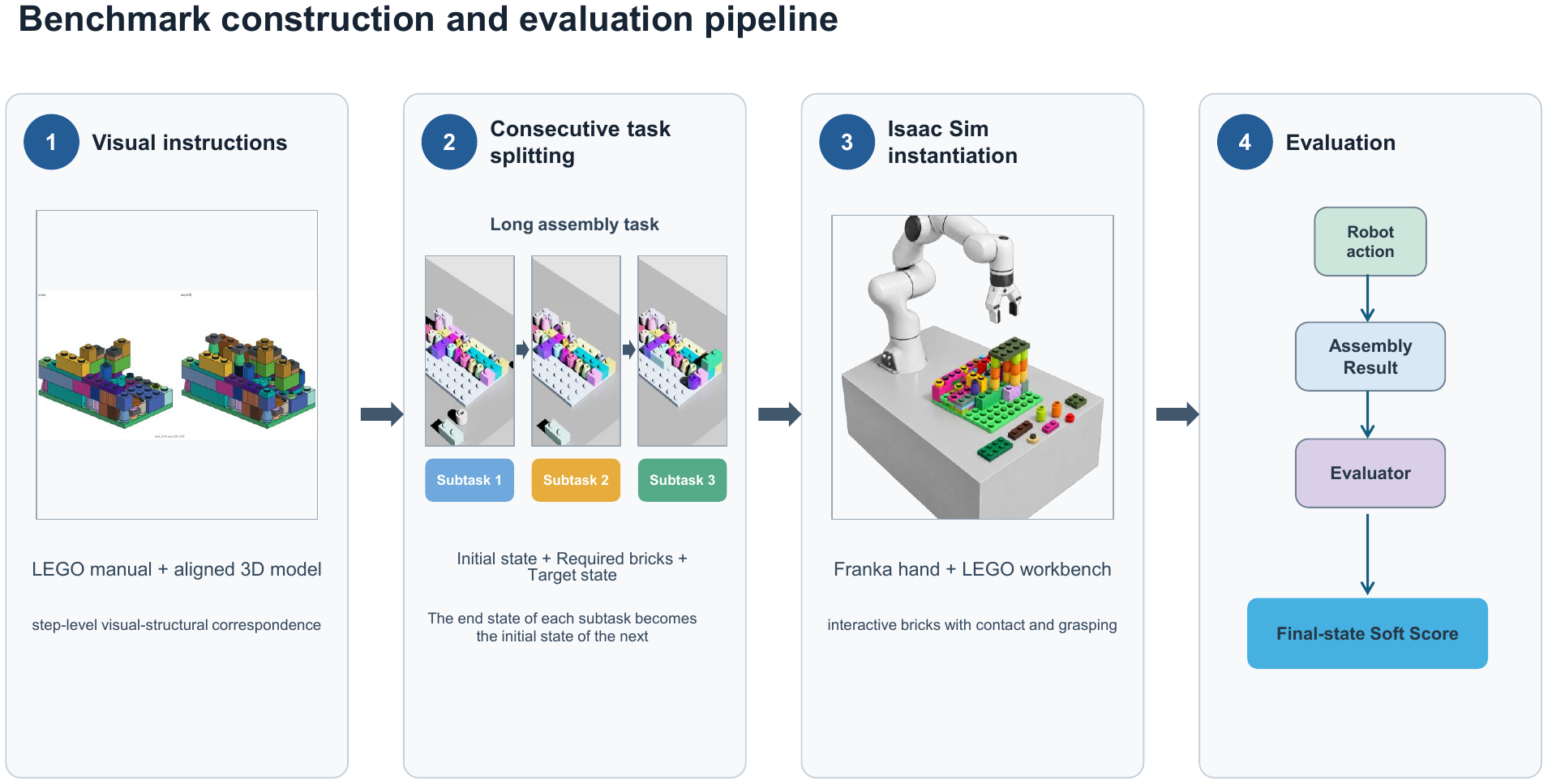}
    \caption{
        Overview of the benchmark construction and evaluation pipeline.
        Visual LEGO instructions are first aligned with their corresponding
        3D structural models. Long construction procedures are then decomposed
        into consecutive, self-contained episodes, each defined by an initial
        state, a set of required bricks, and a target state. These episodes are
        instantiated as physically interactive robotic assembly environments
        in NVIDIA Isaac Sim. During evaluation, a method observes the current
        scene, produces robot actions, and receives updated observations through
        closed-loop interaction. Performance is measured from the resulting
        construction state using the proposed Soft Score.
    }
    \label{fig:benchmark_pipeline}
\end{figure*}

Figure~\ref{fig:benchmark_pipeline} summarizes the benchmark construction and evaluation pipeline. We begin with visual LEGO instructions paired with their corresponding 3D structural models, preserving step-level correspondence between instructional images and executable assembly states. Rather than treating an entire manual as a single episode, we divide each construction procedure into consecutive task segments. Each segment specifies a pre-assembled initial state, the components that must be manipulated, and the corresponding target construction. The resulting tasks are instantiated in NVIDIA Isaac Sim with physical contact, friction, collision, and robot-control interfaces. Evaluated methods interact with the environment in closed loop, while performance is computed from the final physical construction rather than agreement with a reference action trajectory.

We provide a standardized simulation environment with a Franka Panda robotic platform and simulated LEGO bricks. A total of 50 assembly tasks are designed for evaluation, covering easy, medium, and difficult levels of assembly complexity. For each task, agent are given the corresponding LEGO assembly manual and are required to autonomously construct the target structure in the provided simulation environment.

Each benchmark task is instantiated as an interactive robotic assembly episode in NVIDIA Isaac Sim 6.0.1~\cite{pmlr_isaac_sim}, using a Franka Emika Panda manipulator~\cite{gaz2019franka_emika_panda} with a parallel-jaw gripper. Each scene contains a partially assembled LEGO structure and the movable bricks required by the corresponding manual segment. The agent receives top-down RGB observations and robot-state feedback, and controls the arm and gripper through ROS~2~\cite{macenski2022ros_2_robot}. Robot motion is executed with NVIDIA cuMotion/RMPflow, with MoveIt~\cite{chitta2016moveit} supported as an alternative backend.

The environment provides RGB observations from a single planner camera with a top-down view of the LEGO workspace, together with robot joint-state feedback and interfaces for arm and gripper control. Robot execution is integrated through ROS~2 ~\cite{macenski2022ros_2_robot}. By default, robot motion is generated using NVIDIA cuMotion/RMPflow, with MoveIt~\cite{chitta2016moveit} supported as an alternative backend. LEGO bricks are modeled as dynamic rigid bodies with collision geometry, contact dynamics, and friction.

For each episode, the evaluated method receives the visual instruction manual and observations of the initial scene. Depending on the evaluation setting, it operates either without reflection or with closed-loop reflection. In the non-reflective setting, the model generates an executable program directly from the initial inputs, and the program is executed without intermediate model feedback. In the reflective setting, the model receives an updated scene observation after each execution step, assesses the resulting state, and iteratively determines whether to continue, replan, or terminate. Execution ends when the model declares completion or reaches the predefined interaction or wall-clock budget. Ground-truth structural models, object poses, and target placements are used exclusively for evaluation and are not exposed to the evaluated method, except for methods explicitly designated as privileged or oracle baselines.

\section{Assembly Training Dataset}

We provide more than 200,000 LEGO assembly tasks for model training, constructed from multiple sources, including a synthetically generated LEGO 3D release dataset\citep{Machine_Executable_Plan_jiajun}, LDraw-based LEGO 3D models~\citep{ldraw}, and real-world assembly manuals paired with their LOGO 3D structures. Each 3D model is decomposed into step-wise assembly sequences, where the action brick for each step is identified and the corresponding intermediate assembly scene is reconstructed. For every assembly step, we further record the ground-truth state after the action, including the expected brick configuration and structure state. Each task therefore provides the assembly instructions, target structure, step-level action-brick annotations, intermediate scene configurations, and ground-truth assembly states required to instantiate an interactive robotic simulation environment. This structured formulation enables agents to perform sequential LEGO assembly while providing reliable ground-truth references for evaluating assembly planning, manipulation execution, intermediate-state correctness, and final construction quality.

To improve data reliability, the constructed tasks are further reviewed by three human evaluators to verify the consistency of the assembly instructions, step decomposition, action-brick annotations, and ground-truth states. However, the current dataset does not provide reference execution trajectories for every training task. In particular, we do not prescribe a unique sequence of low-level robot motions, 

\subsection{Limitations and Future Directions}

JARVIS provides assembly tasks and target brick poses, but does not specify robot motion trajectories. Mapping these targets to feasible motions requires embodiment-specific inverse kinematics and collision-aware planning. Developing methods that generalize this mapping across robot platforms remains an open direction.

\subsection{Training-Set Composition and Distribution}

\begin{figure*}[t]
    \centering
    \includegraphics[width=\textwidth]{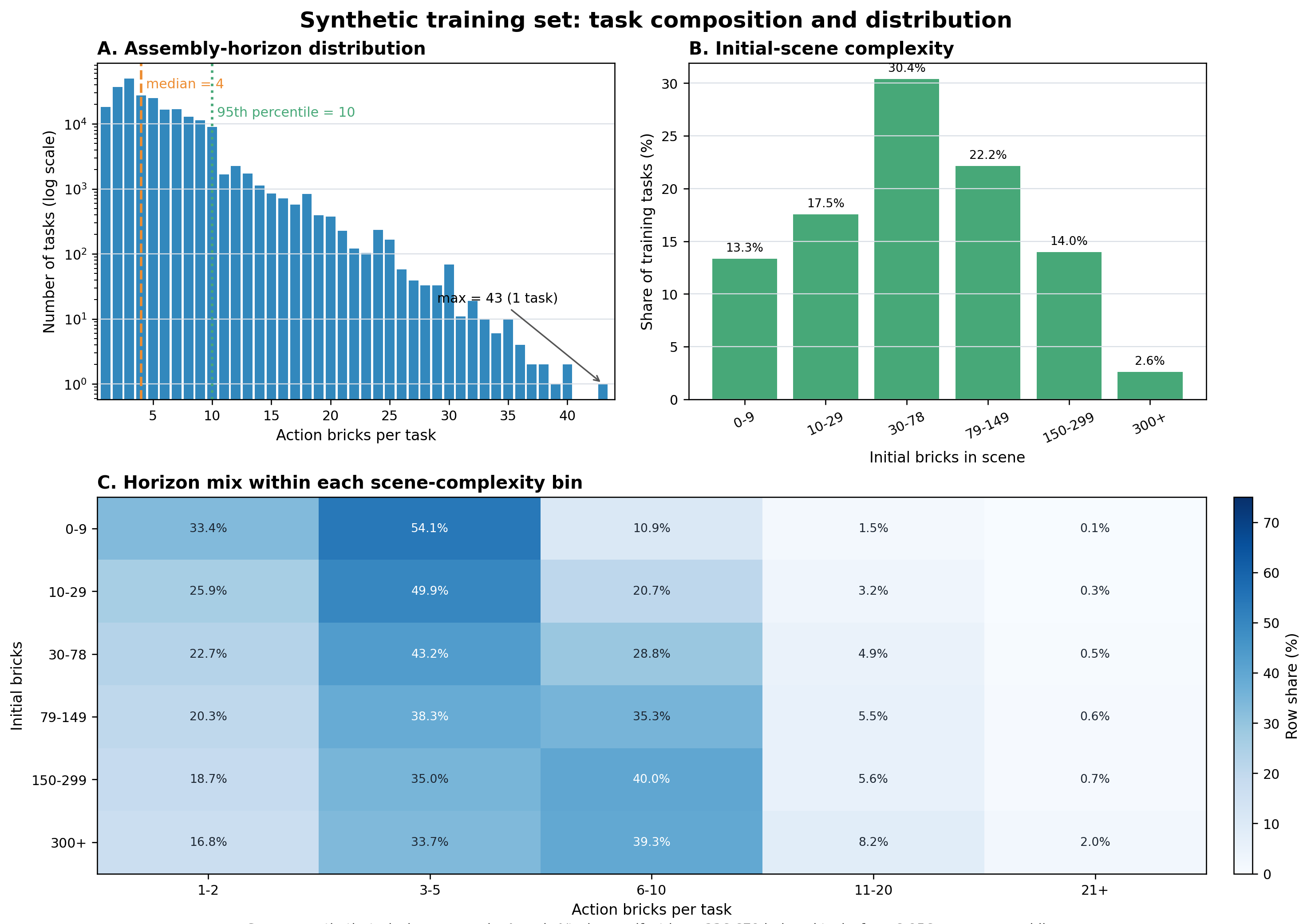}
    \caption{
    Distribution of the LOGO training set.
    \textbf{(a)} Distribution of assembly horizons measured by the number of
    action bricks required in each task.
    \textbf{(b)} Distribution of initial-scene complexity measured by the
    number of bricks present before execution.
    \textbf{(c)} Joint distribution of assembly horizon and initial-scene
    complexity, where each cell reports the horizon composition within the
    corresponding scene-complexity interval.
    }
    \label{fig:training_set_distribution}
\end{figure*}

The training set comprises 236,679 executable tasks derived from 8,987 valid LEGO assemblies, spanning a broad range of assembly states from sparse local structures to densely populated scenes. As shown in Fig.~\ref{fig:training_set_distribution}(a), the distribution of assembly
horizons is strongly right-skewed. Each task requires an average of 4.94 action bricks, with a median of 4, and approximately 95\% of the tasks involve no more than 10 action bricks. Tasks requiring more than 20 action bricks account for only 0.49\% of the dataset, while the longest task contains 43 action bricks. The dataset therefore primarily consists of short- to medium-horizon tasks, while retaining a small number of long-horizon examples that capture more challenging cases involving sequential decision-making, state updates, and error accumulation.

Initial-scene complexity exhibits substantially broader coverage, ranging from 0 to 740 pre existing bricks, with a median of 57. As illustrated in Fig.~\ref{fig:training_set_distribution}(b), tasks containing 30--78 and 79--149 initial bricks constitute the largest groups, accounting for 30.4\% and 22.2\% of the dataset, respectively, whereas scenes containing more than 300 initial bricks account for 2.6\%. Consequently, the training set includes both relatively simple scenes that support the acquisition of basic localization and placement capabilities and highly cluttered scenes that require target identification among numerous candidate bricks and complex
structures.

The joint distribution in Fig.~\ref{fig:training_set_distribution}(c) further reveals only a weak correlation between initial-scene complexity and the number of action bricks (Pearson's $r=0.181$). Thus, perceptual complexity and execution horizon are not tightly coupled: the dataset contains both structurally complex tasks requiring only a small number of actions and visually simpler tasks involving extended placement sequences. This distribution enables models to learn target identification in cluttered environments and state tracking over long-horizon execution as partially independent capabilities, while reducing the likelihood that task difficulty can be inferred from scene size alone. Nevertheless, the training distribution
remains dominated by short-horizon tasks, with long-horizon and extremely dense scenes confined to the distribution tail. 

\subsection{Validation Set and Benchmark}
\label{sec:validation_benchmark_distribution}

\begin{figure*}[t]
    \centering
    \includegraphics[width=\textwidth]{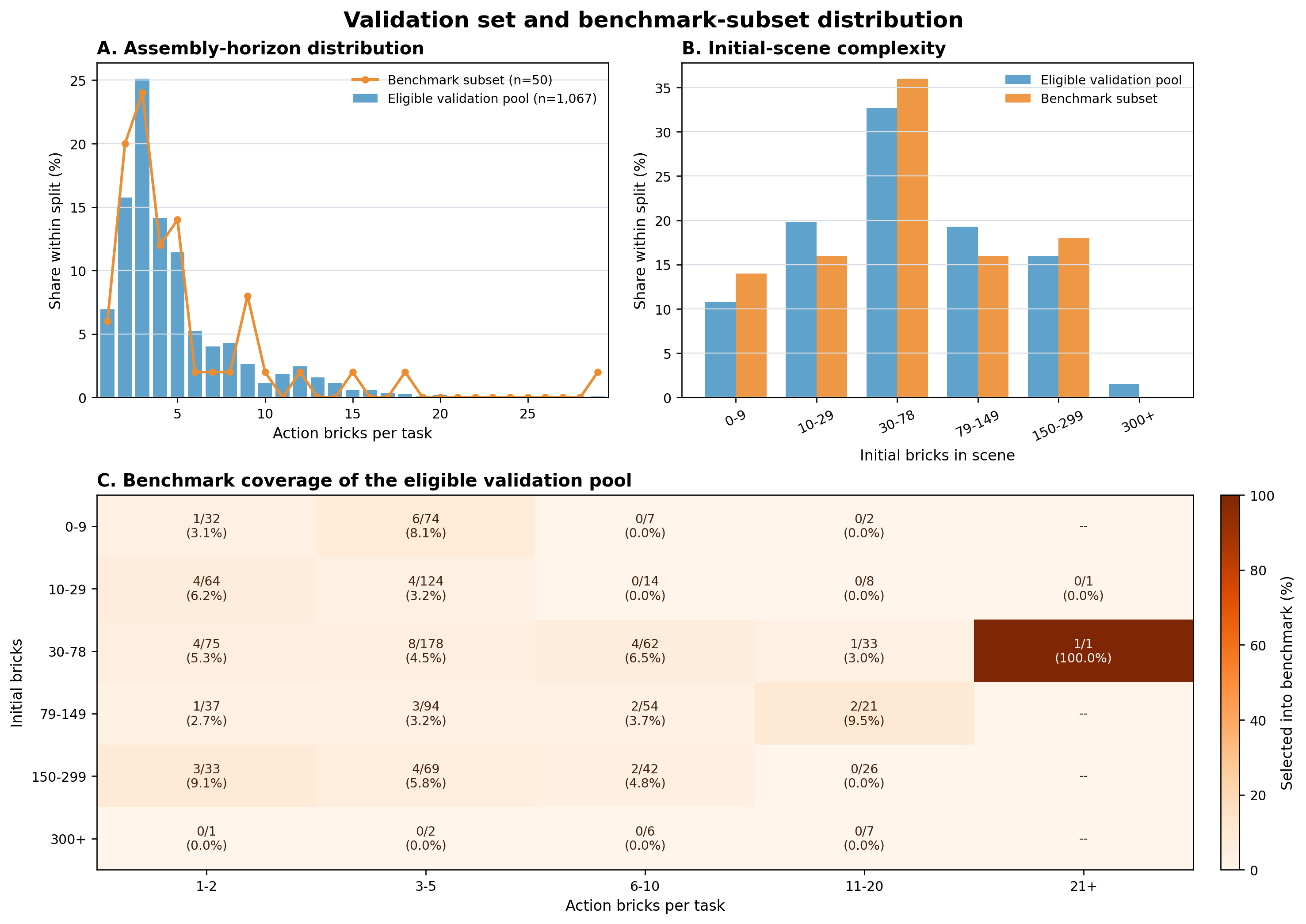}
    \caption{Distributions of the LOGO validation set and the selected \ours    Benchmark. \textbf{(a)} Assembly-horizon distributions measured by the number of action bricks required in each task. \textbf{(b)} Initial-scene complexity distributions measured by the number of bricks present before execution.
    \textbf{(c)} Benchmark coverage over the joint distribution of assembly
    horizon and initial-scene complexity, where each cell reports the number
    of selected Benchmark tasks relative to the number of eligible validation
    tasks in the corresponding interval, together with the selection rate.
    }
    \label{fig:validation_distribution}
\end{figure*}

The original validation set is generated from 48 source LEGO models and contains 1,277 converted tasks. After filtering for scene completeness, geometric validity, and task executability, 1,067 valid candidate tasks remain. These tasks constitute the validation pool from which the evaluation Benchmark is subsequently sampled. The validation set covers a broad range of assembly horizons and initial-scene complexities. As shown in Figure~\ref{fig:validation_distribution}(a), the valid validation tasks require an average of 4.82 action bricks, with a median of 4. Furthermore, 95\% of the tasks require no more than 12 action bricks, while the longest task requires 29 brick-placement actions. The distribution is dominated by short- and medium-horizon tasks: tasks requiring 1--2, 3--5, and 6--10 action bricks account for 22.68\%, 50.70\%, and 17.34\% of the validation pool, respectively. Only two tasks require more than 20 action bricks, corresponding to 0.19\% of the valid validation set.

The number of bricks in the initial scene ranges from 0 to 411, with a mean of 80.44 and a median of 54. As shown in Figure~\ref{fig:validation_distribution}(b), the largest proportion of validation tasks falls within the range of 30--78 initial bricks, while a small number of high-density scenes contain more than 300 bricks. The resulting validation pool therefore covers basic placement tasks, long-horizon assembly tasks, and visually challenging scenes containing substantial clutter.

\paragraph{Benchmark selection.}
Task difficulty is determined using the joint criterion introduced in Section~3. For each candidate task, we compute a normalized initial-scene complexity score $S_i^{\mathrm{scene}}$ from the number of bricks present at the beginning of the task and a normalized assembly-workload score $S_i^{\mathrm{act}}$ from the number of action bricks required to complete the target structure. The overall task difficulty is defined as

\begin{equation}
D_i
=
0.3S_i^{\mathrm{scene}}
+
0.7S_i^{\mathrm{act}}.
\end{equation}

Assembly workload is assigned a larger weight because localization errors, grasp failures, and state-estimation errors may accumulate over consecutive execution steps.

After ranking the 1,067 valid candidate tasks according to $D_i$, we assign the bottom 40\% to the Easy candidate stratum, the middle 30\% to the Medium stratum, and the top 30\% to the Hard stratum. Within each stratum, tasks are sampled at approximately uniform intervals along the ranked difficulty distribution. When multiple candidates have similar difficulty scores, tasks originating from different source LEGO models are prioritized to reduce structural redundancy and increase model diversity. This procedure produces a 50-task Benchmark comprising 20 Easy, 15 Medium, and 15 Hard tasks.

\paragraph{Benchmark distribution.}
The \ours Benchmark contains 50 representative tasks selected from the 1,067 valid validation tasks. Each Benchmark task requires an average of 5.18 action bricks, with a median of 3.5 and a maximum of 29. The initial scenes contain an average of 74.96 bricks, with a median of 58.5 and a maximum of 236. The assembly-horizon distribution remains concentrated in the 3--5 action-brick range, which contains 25 tasks and accounts for 50\% of the Benchmark. At the same time, the Benchmark retains an extreme long-horizon task requiring 29 action bricks. This allows the compact evaluation set to assess error accumulation and long-term state tracking under extended execution sequences. In terms of initial-scene complexity, the Benchmark covers scenes ranging from sparse configurations to highly complex scenes containing 150--299 initial bricks, although no task with more than 300 initial bricks is selected.

\paragraph{Relationship between the validation set and the Benchmark.}
The Benchmark is a representative subset of the valid validation pool rather than an independent data split. Figures~\ref{fig:validation_distribution}(a) and~\ref{fig:validation_distribution}(b) compare their relative distributions with respect to assembly horizon and initial-scene complexity. Overall, the Benchmark preserves the principal characteristics of the validation distribution. In both sets, tasks requiring 3--5 action bricks form the largest group, and initial-scene complexity is most frequently concentrated in the 30--78-brick interval.

Figure~\ref{fig:validation_distribution}(c) further illustrates how the Benchmark covers the joint distribution of initial-scene complexity and assembly horizon. The selected tasks cover most representative combinations of the two factors, while the inclusion of the unique 29-action-brick task strengthens evaluation at the long-horizon tail. Nevertheless, because the Benchmark contains only 50 tasks, several sparsely populated combinations are not represented, particularly the extremely complex region containing more than 300 initial bricks. The purpose of the Benchmark is therefore to cover the principal difficulty gradients of the validation pool under a manageable execution cost, rather than to reproduce every extreme case in the complete validation distribution. Experimental results should consequently be interpreted as a systematic evaluation over the selected representative task range, rather than as an exhaustive measurement of all long-tail cases in the validation set.

\begin{figure}[t]
    \centering
    \includegraphics[width=\linewidth]{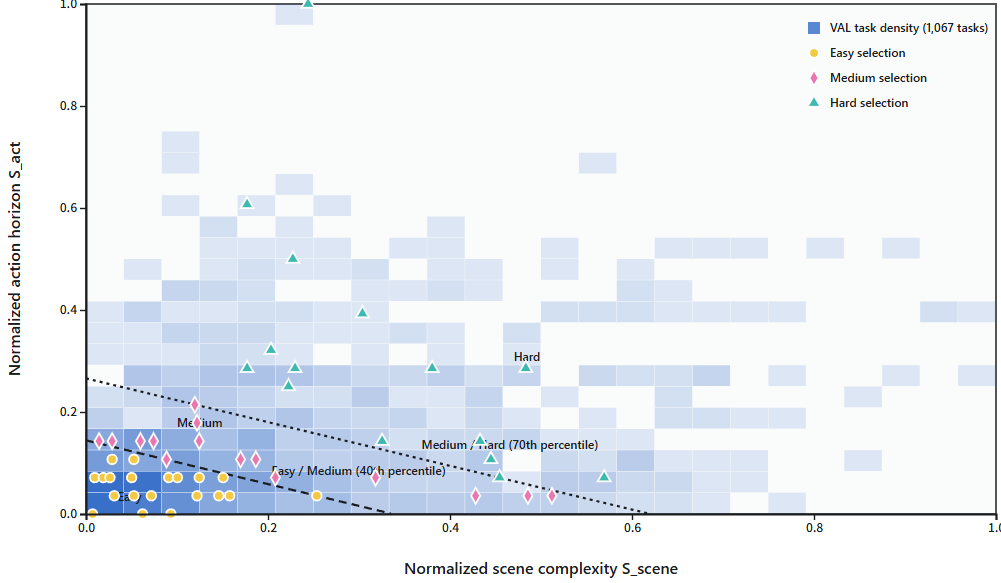}
    \caption{
        Distribution of the 1,067 eligible VAL tasks and the selected
        Benchmark-50 tasks across normalized scene complexity and action
        horizon. For each occupied grid cell, color intensity represents the
        mean normalized difficulty of the tasks within that cell, with lighter
        and darker colors indicating lower and higher difficulty, respectively.
        Empty cells contain no eligible tasks. The overlaid markers denote the
        selected Easy, Medium, and Hard tasks, while the dashed boundaries
        indicate the 40th and 70th percentiles used to define the three
        difficulty strata.
    }
    \label{fig:benchmark_selection_heatmap}
\end{figure}

\begin{figure*}[t]
    \centering
    \includegraphics[width=\textwidth]{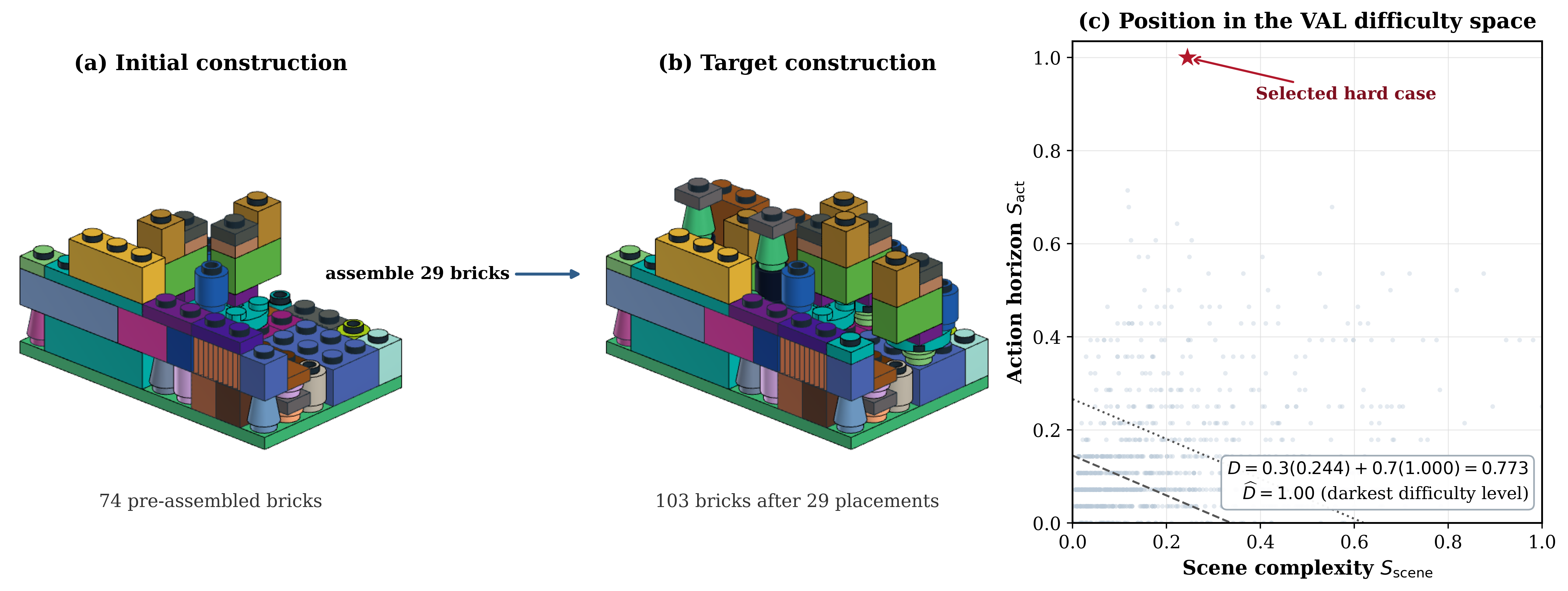}
    \caption{
        A representative Hard assembly task from Benchmark-50. The initial
        structure contains 74 pre-assembled bricks, and the agent must place
        29 additional bricks to obtain the 103-brick target construction.
        The task has $S_{\mathrm{scene}}=0.244$ and
        $S_{\mathrm{act}}=1.000$, yielding
        $D=0.3S_{\mathrm{scene}}+0.7S_{\mathrm{act}}=0.773$ and a normalized
        difficulty of $1.00$ among the 1,067 eligible VAL tasks. Its position
        in the candidate-task distribution is marked in red.
    }
    \label{fig:hard_task_example}
\end{figure*}

Figure~\ref{fig:hard_task_example} presents a representative long-horizon task from Benchmark-50. The episode begins with a partially assembled structure containing 74 bricks and requires the agent to place 29 additional bricks to complete the 103-brick target. Although its initial-scene complexity is moderate relative to the densest validation scenes, its assembly horizon is the longest in the benchmark. The task therefore places substantial demands on instruction understanding, sequential planning, target-brick identification, and state tracking. Because each placement changes the physical scene and constrains subsequent actions, early localization or manipulation errors may propagate throughout the remaining construction sequence. This case exemplifies how the benchmark captures difficulty arising jointly from visual-scene complexity and accumulated long-horizon execution errors.Although the initial construction has only moderate scene complexity, the episode requires the longest normalized action horizon. It is therefore classified as Hard, illustrating how the benchmark captures tasks whose difficulty arises from extended manipulation sequences rather than from scene complexity alone.

\section{Task Execution Pipeline and case study}
\label{sec:task_execution_pipeline}

\subsection{Representative Execution Cases}
\label{sec:representative_execution_cases}

Each task episode follows a unified perception--planning--execution pipeline. The evaluated method first receives a visual assembly manual together with an observation of the current workspace. The manual specifies the target structure and its intended construction procedure, while the scene observation captures the partially assembled structure and the available action bricks. Based on these inputs, the model identifies the relevant bricks, infers the assembly sequence, and determines the next operations to be executed.

The inferred plan is then translated into executable code using the provided API documentation and task-oriented skill library. The generated program can invoke perception, geometric reasoning, grasping, and robot-control functions to localize the selected brick, determine its target placement, and execute the corresponding pick-and-place operation. To maintain a consistent and controlled evaluation interface, all methods must use the provided APIs and manipulation primitives; generated programs are not allowed to redefine these interfaces or directly specify unverified grasp coordinates, inverse-kinematics targets, or gripper forces.

\begin{figure*}[t!]
    \centering
    \includegraphics[width=\textwidth]{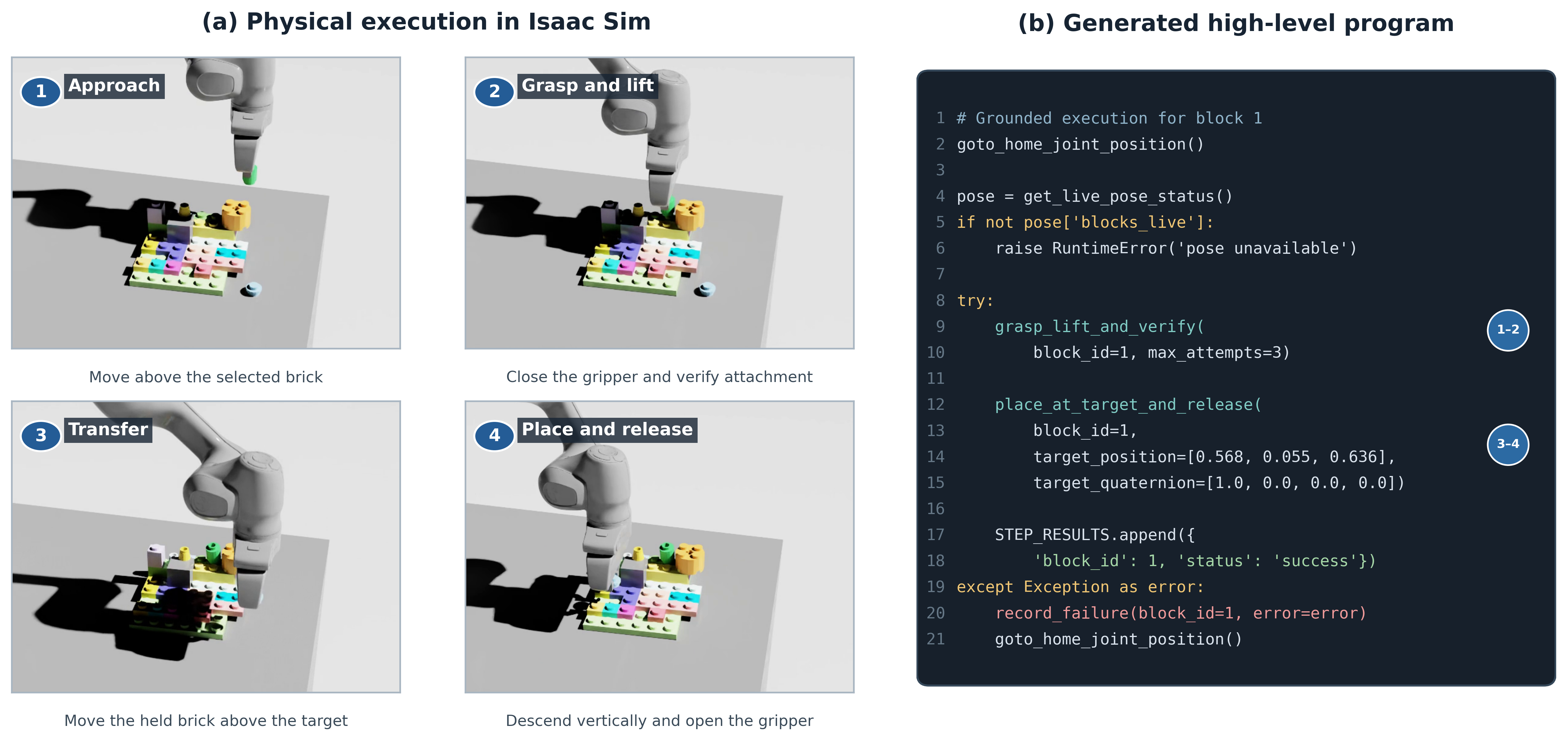}
    \caption{Example of grounded program execution in the interactive assembly environment. The left panel shows four stages from an actual Isaac Sim execution: approaching the selected brick, grasping and lifting it, transferring it above the target, and vertically placing and releasing it. The right panel shows the corresponding high-level program generated for the same operation. The program specifies the selected brick and target pose, while the shared execution primitives handle pose validation, grasp verification, collision-aware motion, and failure recovery.}
    \label{fig:execution_code_example}
\end{figure*}

Figure~\ref{fig:execution_code_example} illustrates how a generated high-level program is grounded into physical robot execution. Given the current scene state and the target pose specified by the plan, the execution layer selects the required brick and invokes shared grasping and placement primitives. The grasping primitive approaches the brick, closes the gripper, verifies attachment, and lifts the brick vertically. The placement primitive transfers the grasped brick above the target pose, performs a vertical descent, and releases it. These primitives internally perform pose validation, grasp verification, collision-aware motion, and failure handling. During execution, updated object poses are continuously recorded, and exceptions are exposed as explicit execution feedback for subsequent agent decisions. 3D descriptor is a compact representation of a brick's geometry and spatial properties, generated using the provided skill tools to support object identification and pose reasoning.

We evaluate two execution modes. In the \emph{non-reflective} mode, the model generates a complete executable program from the initial manual and scene observation, and the program is subsequently executed without further model intervention. In the \emph{reflective} mode, execution proceeds incrementally in a closed loop. After each operation, the model receives an updated workspace observation and execution feedback, assesses whether the resulting state is consistent with the expected assembly progress, and decides whether to continue, revise the plan, perform a corrective action, or terminate. This produces a closed-loop process of observation, reasoning, code generation, execution, and state reassessment.

\subsection{API interface and reflective}
The provided API documentation defines a constrained set of callable skills for scene observation, object grounding, geometric reasoning, point-cloud-based grasp estimation, motion planning, gripper control, and pick-and-place execution. Each function specifies its required inputs, returned observations, and execution constraints, allowing the model to compose robot programs without directly accessing low-level simulator state or issuing unverified control commands. In reflective execution, the model invokes these APIs incrementally rather than generating the entire action sequence at once. After each pick-and-place operation, it receives an updated camera observation and structured execution feedback, compares the observed state with the expected assembly progress, and determines whether the previous action succeeded. The model then either proceeds to the next brick, regenerates the affected portion of the program, performs a corrective action, or terminates the episode. 

\subsection{Execution example}

We provide representative examples of actual robot execution to analyze how different backbone models behave throughout the assembly process and how differences in their generated actions affect the final construction quality. Specifically, we compare the no-reflection execution trajectories of Gemini-3-Pro, Claude Opus-4.5, and DeepSeek-V4.1 on three representative tasks. Each example presents four key stages---the initial scene, approach and grasp, transfer, and final placement---together with the resulting soft score. These cases allow us to distinguish basic execution feasibility from placement quality and illustrate how positional, vertical, and rotational errors lead to different scores even when all models generate complete and executable pick-and-place sequences.

\begin{figure*}[t]
    \centering
    \includegraphics[width=\textwidth]
    {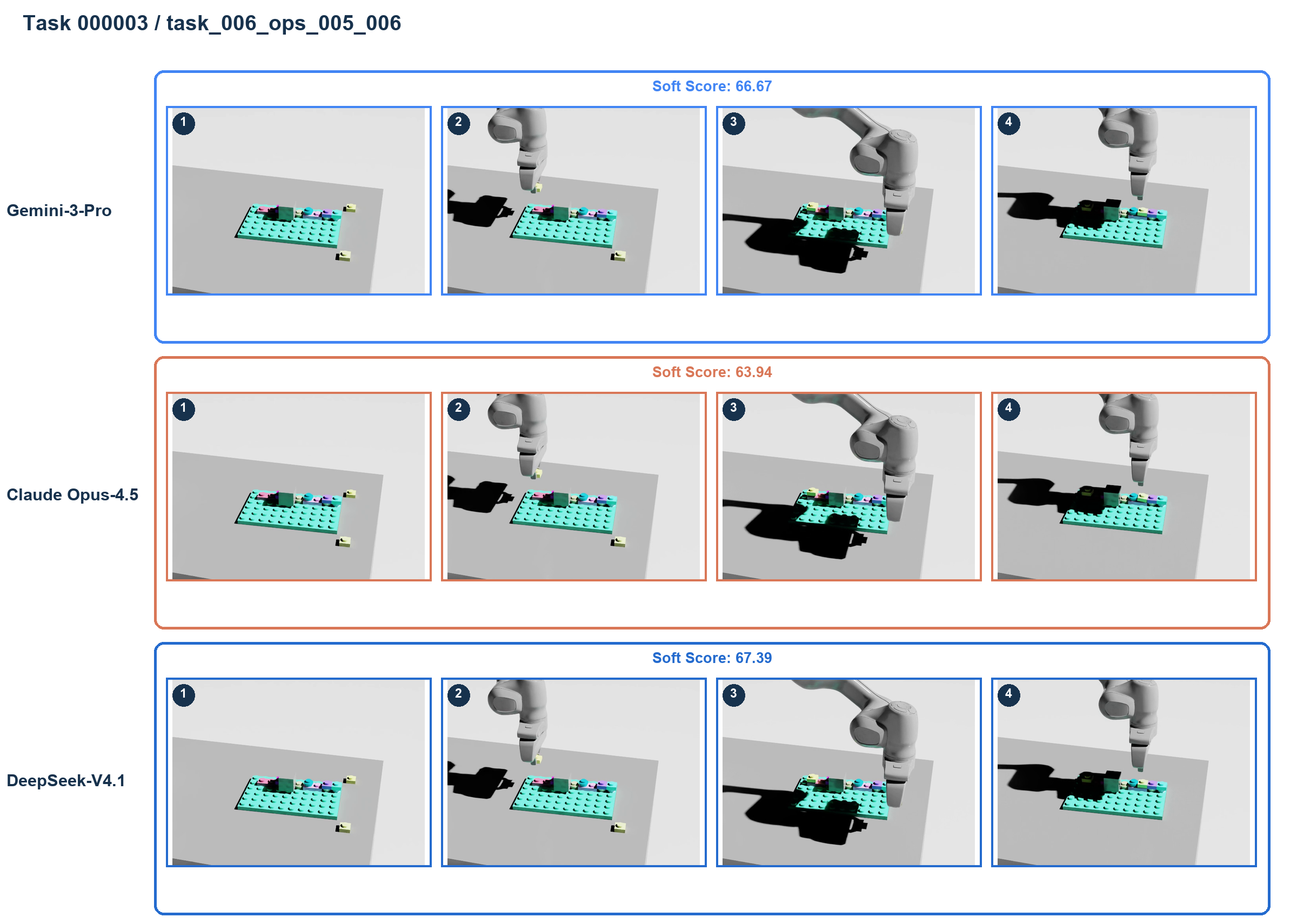}
    \caption{
        No-reflection execution comparison on Task 000003.
        The four frames in each row show the initial scene, approach and grasp,
        transfer, and final placement generated by the corresponding model.
        Gemini-3-Pro, Claude Opus-4.5, and DeepSeek-V4.1 obtain soft scores of
        $66.67$, $63.94$, and $67.39$, respectively.
        All three models complete the manipulation sequence, with the remaining
        performance differences primarily reflecting final placement accuracy.
    }
    \label{fig:execution_task_000003}
\end{figure*}

\paragraph{Task 000003.}
Figure~\ref{fig:execution_task_000003} compares the no-reflection execution trajectories of Gemini-3-Pro, Claude Opus-4.5, and DeepSeek-V4.1 on Task 000003. All three models successfully identify the action brick and generate a complete sequence covering approach, grasping, transfer, and placement. Their final results are visually similar, producing soft scores of $66.67$, $63.94$, and $67.39$, respectively. DeepSeek-V4.1 achieves the highest score, followed closely by Gemini-3-Pro, whereas Claude Opus-4.5 exhibits slightly larger residual placement error. The small score differences indicate that all three models perform reliably on this task and that their remaining differences primarily arise from final pose accuracy rather than execution feasibility.

\begin{figure*}[t]
    \centering
    \includegraphics[width=\textwidth]
    {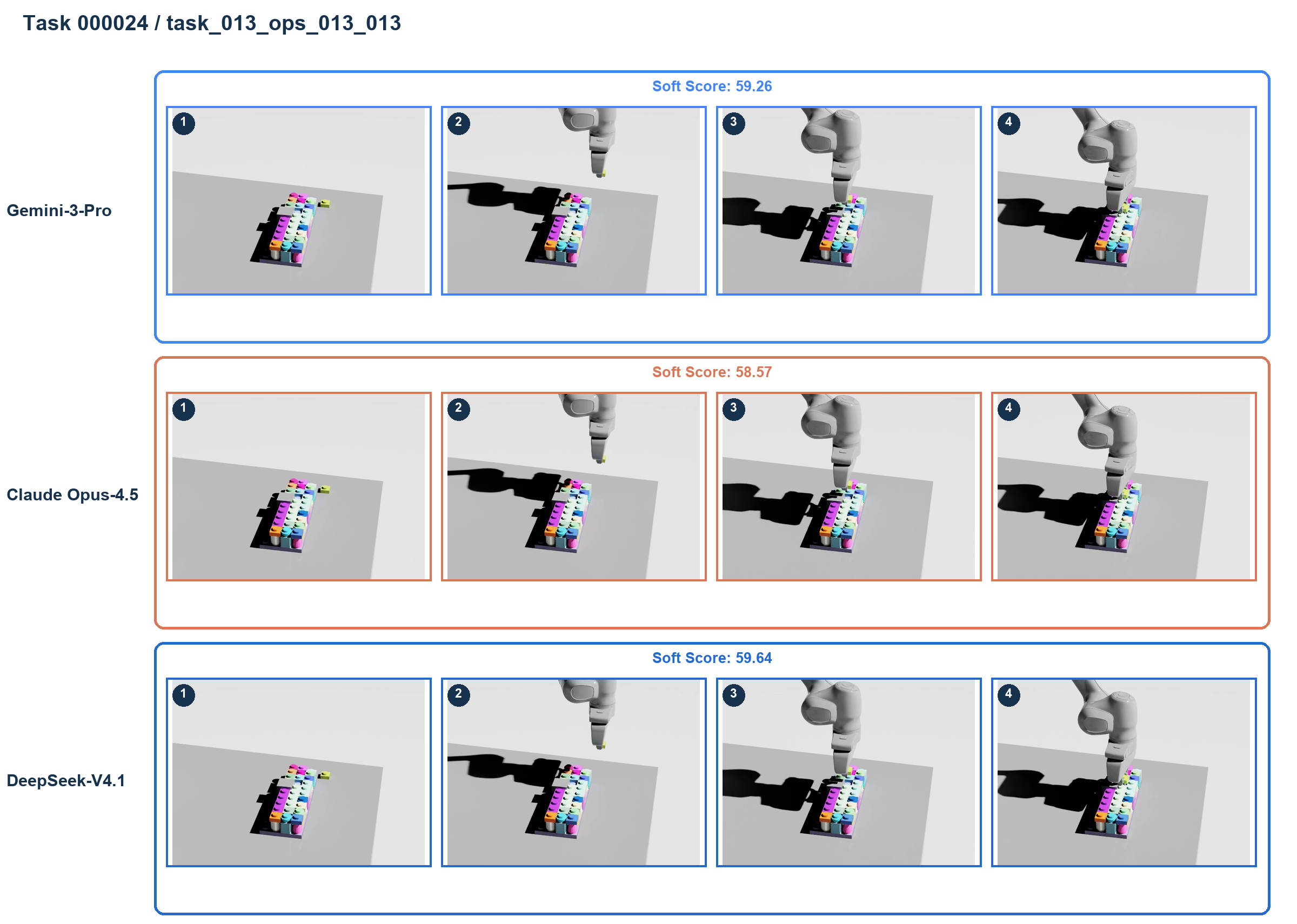}
    \caption{
        No-reflection execution comparison on Task 000024.
        Each row presents four representative stages of execution for one
        backbone model.
        Gemini-3-Pro, Claude Opus-4.5, and DeepSeek-V4.1 achieve soft scores of
        $59.26$, $58.57$, and $59.64$, respectively.
        Their closely grouped scores indicate similar execution quality, with
        only minor differences in the final placement pose.
    }
    \label{fig:execution_task_000024}
\end{figure*}

\paragraph{Task 000024.}
Figure~\ref{fig:execution_task_000024} presents the corresponding comparison on Task 000024. Despite the different backbone models, the three executions follow similar manipulation trajectories and produce comparable final assembly states. Gemini-3-Pro obtains a soft score of $59.26$, Claude Opus-4.5 obtains $58.57$, and DeepSeek-V4.1 obtains $59.64$. The narrow score range suggests that this task is solved with similar effectiveness by all three models. DeepSeek-V4.1 provides the highest-scoring placement, although the margin over Gemini-3-Pro and Claude Opus-4.5 is small. Thus, the observed variation is mainly attributable to minor positional, vertical, or rotational deviations in the final brick pose.

\begin{figure*}[t]
    \centering
    \includegraphics[width=\textwidth]
    {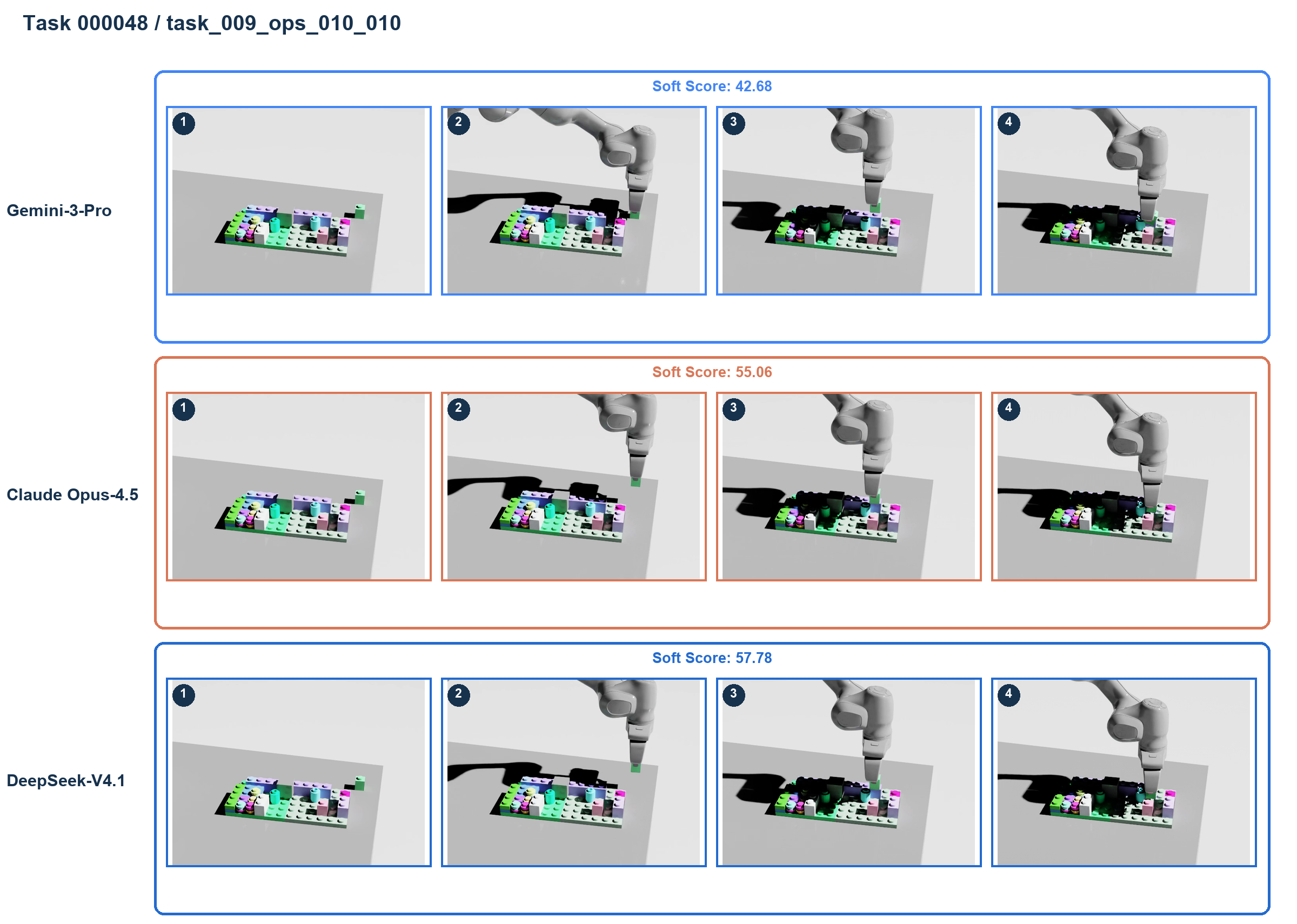}
    \caption{
        No-reflection execution comparison on Task 000048.
        Gemini-3-Pro, Claude Opus-4.5, and DeepSeek-V4.1 obtain soft scores of
        $42.68$, $55.06$, and $57.78$, respectively.
        All three models generate executable manipulation sequences, but their
        final placement accuracy differs substantially.
        DeepSeek-V4.1 produces the highest-quality final state, whereas the
        larger residual pose error of Gemini-3-Pro results in a lower soft score.
    }
    \label{fig:execution_task_000048}
\end{figure*}

\paragraph{Task 000048.}
Task 000048 produces a clearer difference among the three backbone models, as shown in Fig.~\ref{fig:execution_task_000048}. Although all models generate executable pick-and-place sequences, their final placement quality differs substantially. Gemini-3-Pro obtains a soft score of $42.68$, compared with $55.06$ for Claude Opus-4.5 and $57.78$ for DeepSeek-V4.1. The lower Gemini-3-Pro score indicates a larger discrepancy between the achieved and target poses, despite the successful completion of the basic manipulation sequence. Claude Opus-4.5 produces a more accurate final state, while DeepSeek-V4.1 achieves the best alignment among the three models. This case illustrates that successful code execution alone does not guarantee high assembly quality: small errors introduced during grasping, transfer, or release can lead to meaningful differences in the structure-aware soft score.

\begin{figure*}[t]
    \centering
    \includegraphics[width=\textwidth]{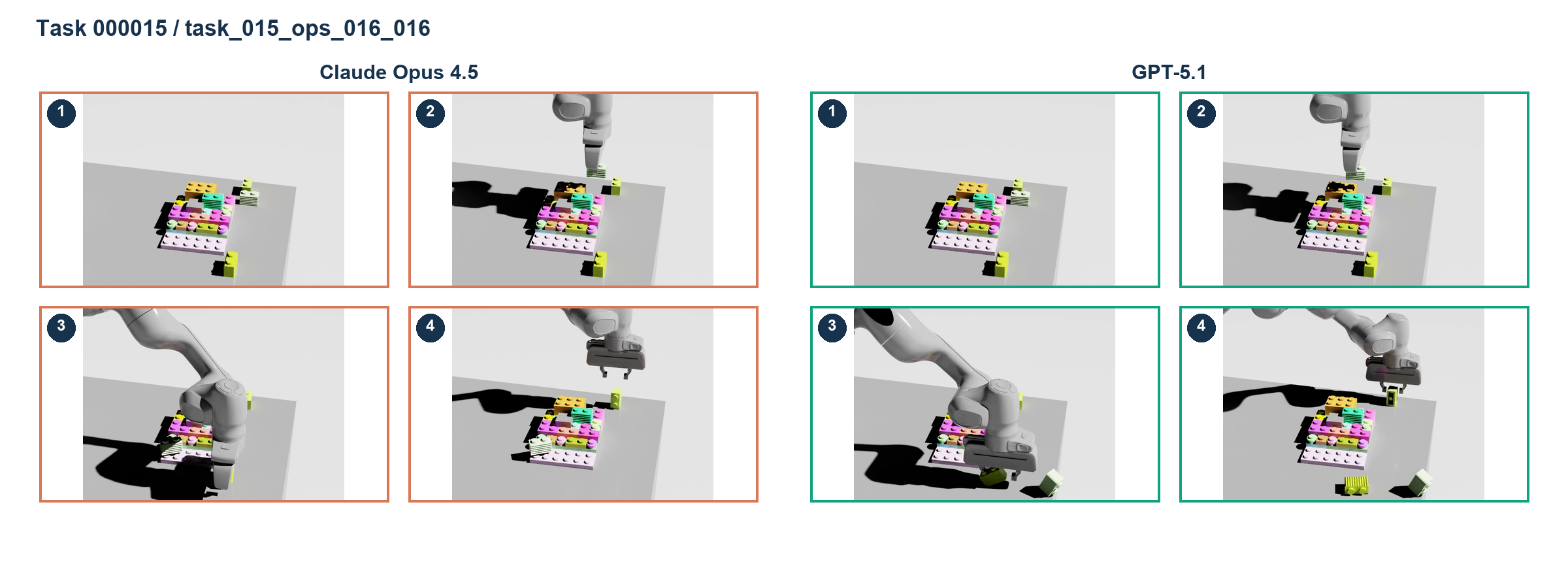}
    \caption{
        Comparison of non-reflective execution by Claude Opus-4.5 and GPT-5.1 on the same low-scoring task.
        The numbered frames show representative stages from the initial scene through grasping and transfer to the final state.
        Although both models generate executable manipulation sequences, unintended collisions with surrounding LEGO bricks disturb the existing structure and cause bricks to shift or scatter.
        Without visual feedback and corrective replanning, the resulting scene deviation persists until termination, yielding soft scores of $0.43$ for Claude Opus-4.5 and $0.00$ for GPT-5.1.
    }
    \label{fig:low_score_execution_comparison}
\end{figure*}

\paragraph{Case 1: Collision-Induced Failure in Non-Reflective Execution.}
\label{para:collision_failure}
Moreover, from the perspective of failure modes, an important issue is not the model's inability to generate basic pick-and-place operations, but unintended collisions between the robot arm or the grasped brick and surrounding LEGO bricks during execution. As illustrated in Fig.~\ref{fig:low_score_execution_comparison}, both Claude Opus-4.5 and GPT-5.1 successfully approach and manipulate the selected brick in the non-reflective setting. However, collisions occurring during transfer and placement disturb the surrounding structure, causing neighboring bricks to shift, topple, or scatter. Consequently, the actual workspace state progressively deviates from that assumed by the initially generated program. Because no updated visual feedback is provided for replanning or corrective execution, these disturbances propagate through the remaining actions and ultimately result in assembly failure.

\begin{figure*}[t]
    \centering
    \includegraphics[width=\textwidth]{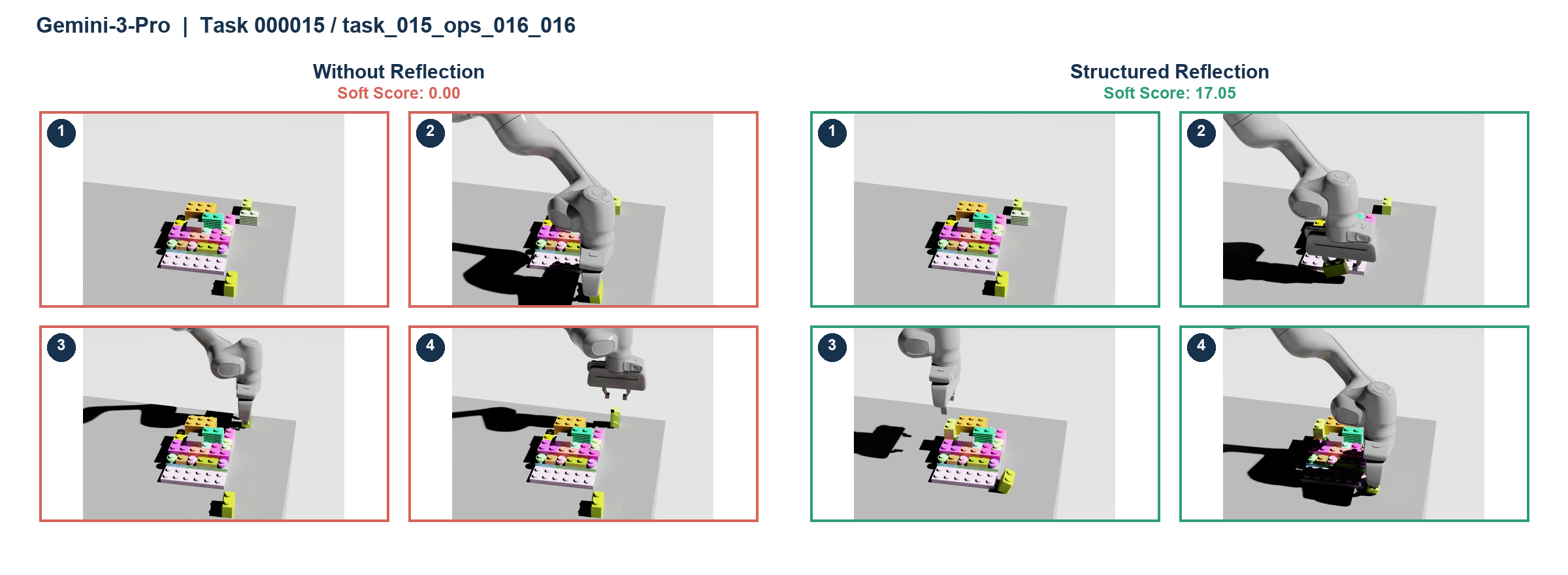}
    \caption{
        Comparison of Gemini-3-Pro execution with and without structured reflection on the same assembly task.
        Without reflection, an unsuccessful or misaligned grasp remains undetected and propagates to subsequent actions, producing a soft score of $0.00$.
        With structured reflection, the model uses updated visual observations to reassess the workspace state and generate corrective actions, improving the soft score to $17.05$.
        The numbered frames show representative stages from the initial scene to the final assembly state.
    }
    \label{fig:reflection_before_after}
\end{figure*}

\paragraph{Case 2: Recovery from Grasp Failure through Reflection.}
\label{para:reflection_recovery}
When a grasp cannot be executed successfully, reflection enables the model to inspect the updated visual observation and determine whether the target brick has been correctly grasped and whether the resulting scene is consistent with the expected state. If the brick remains at its initial location, the grasp is misaligned, or the operation disturbs the surrounding structure, the model can re-identify the target, adjust the grasp pose or approach trajectory, and generate a corrective action instead of continuing with a plan based on an incorrect state. As illustrated in Fig.~\ref{fig:reflection_before_after}, the grasp failure propagates to the subsequent placement stage without reflection, resulting in a soft score of $0.00$. Structured reflection partially recovers the assembly process through state reassessment and corrective execution, increasing the soft score to $17.05$. This example demonstrates that reflection primarily limits the propagation of local grasping errors, although it does not guarantee complete task recovery.

\begin{figure*}[t]
    \centering
    \includegraphics[width=\textwidth]{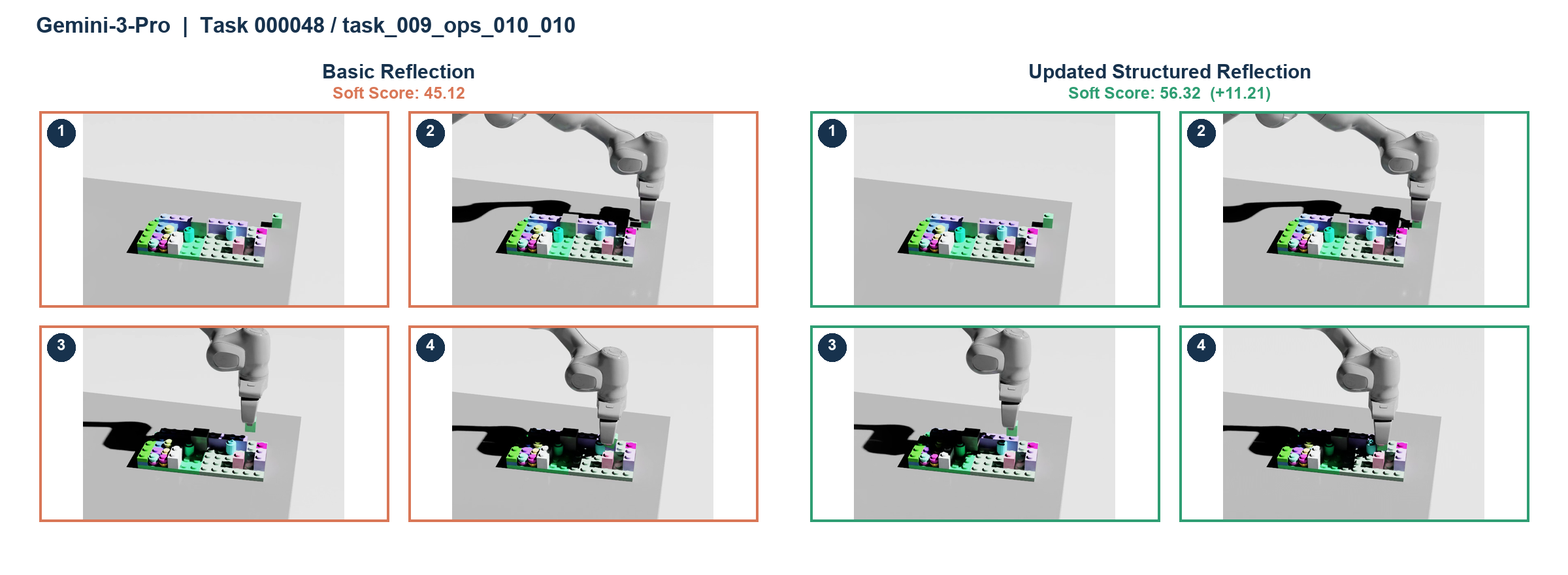}
    \caption{
        Comparison between \textit{Standard Reflection} and
        \textit{Reflection + Prior-Task Experience Summary} on the same
        Gemini-3-Pro assembly task.
        The numbered frames show representative stages of the two executions.
        Incorporating prior-task experience produces a more accurate placement,
        particularly in orientation, increasing the soft score from $45.12$ to
        $56.32$.
        Although the pose error is reduced, neither setting satisfies the strict
        completion criterion.
    }
    \label{fig:prior_experience_comparison}
\end{figure*}

\clearpage
\paragraph{Case 3: Improved Reflection with Prior-Task Experience.}
\label{para:experience_improvement}
As an example of improved execution, we compare the experience-augmented method, corresponding to the \textit{Reflection + Prior-Task Experience Summary} setting in Table~5, with \textit{Standard Reflection} on the same task. Incorporating prior-task experience primarily improves placement-pose accuracy by helping the model identify positional, vertical, and orientational deviations and translate them into better-constrained corrective actions. In particular, the orientation score increases from $24.57$ to $80.63$, while the position score improves from $96.20$ to $99.72$, raising the overall soft score from $45.12$ to $56.32$. Both settings achieve an execution-validity score and a stability score of $100\%$, indicating that the improvement arises from the quality of the corrected placement rather than program executability. Nevertheless, the completion score remains zero, showing that prior-task experience substantially reduces the pose error but does not fully recover the intended structural connection in this episode, as illustrated in Fig.~\ref{fig:prior_experience_comparison}.

\begin{table}[t]
\centering
\caption{
Effective-action performance under ground-truth-guided assembly. Scores are averaged over effective actions against brick-specific ground truth, without execution-time penalties.
}
\label{tab:overall_effective_action_score}
\begin{tabular}{lcc}
\toprule
Model & Effective actions & Average score \\
\midrule

Gemini 3 Pro      & 150          & \textbf{24.25} \\
Claude Opus 4.5   & 138          & 19.68 \\
Claude Haiku 4.5  & 112          & 17.66 \\
DeepSeek-V4.1     & 142          & 22.66 \\
Qwen3-VL-235B     & 127          & 19.23 \\
Kimi-K2           & 112          & 17.12 \\
GPT-4o            & 86           & 13.20 \\
GPT-5.1           & 146          & 19.76 \\
GPT-OSS-120B      & \textbf{187} & 23.76 \\

\bottomrule
\end{tabular}
\end{table}

\section{Execution-stage analysis.}

To further isolate execution quality from planning and grounding errors, we evaluate effective actions under ground-truth-guided assembly, as shown in Table~\ref{tab:overall_effective_action_score}. For each target brick, the correct brick-specific assembly specification is provided, allowing the model to focus on executing the corresponding manipulation action rather than inferring the next assembly step. The reported score is averaged over effective actions and therefore measures conditional execution quality when a valid assembly target is available.

The results reveal that execution capability and action coverage are distinct factors. Gemini-3-Pro achieves the highest average effective-action score of 24.25 over 150 actions, indicating the strongest per-action execution quality among the evaluated models. In contrast, GPT-OSS-120B produces the largest number of effective actions (187) while achieving a comparable average score of 23.76. DeepSeek-V4.1 also obtains a relatively high average score of 22.66 over 142 effective actions. These results show that some models can execute individual assembly actions reasonably well once the correct target and placement specification are given, even though their full-pipeline assembly performance remains substantially lower.

This gap helps explain why intermediate planning and grounding metrics do not directly translate into successful assembly. In the full pipeline, the agent must repeatedly select the correct construction step, localize the target brick, generate a feasible manipulation command, execute the action, and continue from the resulting physical state. Errors at any stage can propagate to subsequent steps. Consequently, a model with strong conditional execution quality may still achieve a low final assembly score if it fails to generate sufficient valid actions or makes planning errors earlier in the sequence. Conversely, a model that covers more assembly steps may still accumulate larger placement or manipulation errors. This motivates evaluating the final assembled state in addition to planning, grounding, and single-action execution metrics.

\end{document}